\documentclass{article}

\usepackage{arxiv}

\usepackage[utf8]{inputenc} 
\usepackage[T1]{fontenc}    
\usepackage{hyperref}       
\usepackage{url}            
\usepackage{booktabs}       
\usepackage{amsfonts}       
\usepackage{nicefrac}       
\usepackage{microtype}      
\usepackage{lipsum}		
\usepackage{graphicx}
\usepackage{natbib}
\usepackage{doi}

\usepackage{amssymb}
\usepackage{varwidth}
\usepackage{amsmath}

\makeatletter
\long\def\@maketablecaption#1#2{\@tablecaptionsize
    \global \@minipagefalse
    \hbox to \hsize{\parbox[t]{\hsize}{\centering #1 \\ #2}}}

\makeatother
\newcommand{\bunderline}[2][4]{\underline{#2\mkern-#1mu}\mkern#1mu}

\newcommand{\nunder}[2][5]{\mathrlap{\mkern\the\numexpr#1/2mu\relax\underline{\phantom{\mathrm{#2}\mkern-#1mu}}}#2}

\makeatletter
\newenvironment{tablehere}
{\def\@captype{table}}
{
	
}
\newenvironment{figurehere}
{\def\@captype{figure}}
{}
\makeatother
\makeatletter
\newsavebox\tboxa
\newsavebox\tboxb
\newlength\tdima

\newcommand*{\oversymb}{\mathpalette\@oversymb}

\newcommand*{\@oversymb}[2]{%
    \sbox{\tboxa}{$\m@th#1\mathrm{#2}$}%
    \setbox\tboxb\null%
    \ht\tboxb\ht\tboxa%
    \dp\tboxb\dp\tboxa%
    \wd\tboxb\wd\tboxa%
    \sbox{\tboxa}{$\m@th#1{#2}$}%
    \setlength\tdima{\the\wd\tboxa}%
    \addtolength\tdima{-\the\wd\tboxb}%
    \sbox{\tboxb}{$\m@th#1\hskip\tdima\overline{\xusebox{\tboxb}}$}%
    \rlap{\usebox\tboxb}{\usebox\tboxa}}

\newcommand*{\xusebox}[1]{\mathord{{\usebox{#1}}}}

\DeclareMathOperator*{\argmax}{arg\,max}
\title{A Hybrid Multilayer Extreme Learning Machine for Image Classification with an Application to Quadcopters}

\author{ \href{https://orcid.org/0000-0000-0000-0000}{\includegraphics[scale=0.06]{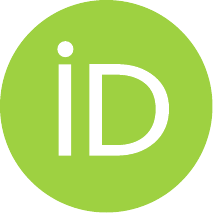}\hspace{1mm}Rolando~A.~Hernandez-Hernandez} \\
	Laboratory of Submarine Robotics (LSR)\\
	Center for Engineering and Industrial Development (CIDESI)\\
	76125, Qro. M\'{e}xico \\
	\texttt{r.hernandez@posgrado.cidesi.edu.mx} \\
	\And
	\href{https://orcid.org/0000-0000-0000-0000}{\includegraphics[scale=0.06]{orcid.pdf}\hspace{1mm}Adrian~Rubio-Solis} \\
	Hamlyn Centre for Robotic
Surgery\\
	Imperial College London\\
	 London, SW7 2AZ, UK\\
	\texttt{arubioso@ic.ac.uk} \\
}

\renewcommand{\shorttitle}{\textit{HML-ELM for image classification with an application to quadcopters} }

\hypersetup{
pdftitle={A template for the arxiv style},
pdfsubject={q-bio.NC, q-bio.QM},
pdfauthor={David S.~Hippocampus, Elias D.~Striatum},
pdfkeywords={First keyword, Second keyword, More},
}

\begin{document}
\maketitle

\begin{abstract}
	Multilayer Extreme Learning Machine (ML-ELM) and its variants have proven to be an effective technique for the classification of different natural signals such as audio, video, acoustic and images. In this paper, a Hybrid Multilayer Extreme Learning Machine (HML-ELM) that is based on ELM-based autoencoder (ELM-AE) and an Interval Type-2 fuzzy Logic theory is suggested for active image classification and applied to Unmanned Aerial Vehicles (UAVs). The proposed methodology is a hierarchical ELM learning framework that consists of two main phases: 1) self-taught feature extraction and 2) supervised feature classification. First, unsupervised multilayer feature encoding is achieved by stacking a number of ELM-AEs, in which input data is projected into a number of high-level representations. At the second phase, the final features are classified using a novel Simplified Interval Type-2 Fuzzy ELM (SIT2-FELM) with a fast output reduction layer based on the SC algorithm; an improved version of the algorithm Center of Sets Type Reducer without Sorting Requirement (COSTRWSR). To validate the efficiency of the HML-ELM, two types of experiments for the classification of images are suggested. First, the HML-ELM is applied to solve a number of benchmark problems for image classification. Secondly, a number of real experiments to the active classification and transport of four different objects between two predefined locations using a UAV is implemented. Experiments demonstrate that the proposed HML-ELM delivers a superior efficiency compared to other similar methodologies such as ML-ELM, Multilayer Fuzzy Extreme Learning Machine (ML-FELM) and ELM.
\end{abstract}

\keywords{Multilayer Extreme Learning Machine \and Interval Type-2 Fuzzy Logic Systems \and Unmanned Aerial Vehicles \and Active image classification.  }

\section{Introduction}

During the last decade, Extreme Learning Machine (ELM) has been studied by many researches in theory and applications \cite{park2019convolutional, rubio2020multilayer, tang2016extreme, vong2018empirical,qing2020deep, wang2017encrypted}. ELM was originally introduced as a unified learning framework   for generalized single-hidden layer feedforward Networks (SLFNs), including
but not limited to sigmoid networks, radial basis function
networks \cite{rubio2018evolutionary, hernandez2020multilayer}, convolutional neural networks \cite{park2019convolutional}, type-1 and type-2 fuzzy inference systems \cite{rubio2020multilayer}, high-order networks \cite{christou2019hybrid}, wavelet networks \cite{ding2016wavelet}, deep neural networks \cite{ding2015deep, huang2006extreme}, etc. According to traditional Neural Networks (NNs) theory, a SLFN is a universal approximator when all its parameters, i.e. hidden and input weights as well as biases are allowed to be adjustable \cite{yang2015multilayer}.  ELM is a fully random learning method that differs from conventional theories for NNs. In ELM-based methods, hidden layer parameters of NNs are randomly generated and need not to be optimised during training (including input weights and hidden layer biases), while the output weights are analytically determined by using the generalised inverse method \cite{huang2006extreme}. In a significant number of applications and developments, it has been demonstrated that ELM delivers a superior performance than other literature techniques such as SVM/LS-SVM \cite{yang2015multilayer}. This includes  the solution of problems such as sparse and Bayesian learning  for classification \cite{luo2013sparse}, computer vision \cite{tang2016extreme}, data clustering \cite{huang2018clustering}, regression challenges \cite{rubio2018evolutionary}, high-dimensional and large data sets \cite{cao2015extreme,cao2014protein}, etc. However, in many real world applications, the aforementioned solutions still face issues when dealing with natural signals such as video, audio, voice recognition, etc \cite{tang2016extreme}. Many of these ELM applications mainly involve the classification of signals, where a process for feature learning is often conducted before classifications \cite{tang2016extreme}. As pointed out in \cite{tang2016extreme}, multilayer solutions are frequently required. Within this context, a number of multilayer ELM (ML-ELM) solutions have been suggested, in which two main steps are frequently needed. First, a step for feature extraction based on ELM Autoencoders (ELM-AEs) where a high level of representation of input data is obtained. In the second step, these features are finally classified using traditional ELM. For example in \cite{tang2016extreme},  a hierarchical ELM (H-ELM) for the effective training of ML Perceptron neural networks was suggested. The proposed H-ELM consists of two main stages: a) a unsupervised multilayer feature encoding and b) supervised feature classification. The H-ELM was successfully applied to solve three different problems in the field of computer vision, namely, a) car detection, b) gesture recognition, and c) online incremental tracking. In H-ELM, original inputs are decomposed into multiple hidden layers and the outputs of previous layers are used as the input of the current one. At each layer in the process for feature extraction, each module can be viewed as an independent process where features are first scattered by a random matrix, and traditional ELM is applied for final decision.

In other fields such as medicine, ML-ELM structures have also been successfully developed. For example, in \cite{rubio2020multilayer}, a ML-ELM based on the concept of Interval Type-2 Fuzzy Logic was developed for the recognition of walking activities and gait events. The proposed ML structure called Multilayer Interval-2 Fuzzy ELM (ML-IT2FELM) was applied to analyse movement data collected from various disabled people and used for the classification of time series. The proposed ML-IT2FELM was introduced as a hierarchical learning framework that consists of two main phases. First a number of multiple IT2 Fuzzy Autoencoders (IT2-FAEs) for feature extraction was implemented, followed by a final classification layer based on IT2-FELM. Compared to H-ELM, ML-IT2FELM incorporates IT2 fuzzy sets in each hidden layer to provide a better treatment of noisy data usually found in the classification of signals collected by wearable sensors. In \cite{rubio2020multilayer}, unlike typical ELM-AEs, two major advantages are provided, first each FAE in the ML-IT2FEM does not need to transform input data into a random feature ELM space. Secondly, an improved generalisation performance in the presence of uncertainty is achieved. At the ML-IT2FELM, authors employed a direct defuzzification method based on the Nie-Tan approach (NT) for feature extraction while a Karnik-Mendel algorithm (KM) was applied as the output layer of the final classification module. On the one hand, NT algorithms are a first-order approximation of KM algorithms \cite{nie2008towards} while KM usually involves an iterative and permutation process for the calculation of the output weights (or called consequent parts in fuzzy logic) in the final classification layer. The optimisation of IT2 Fuzzy Logic Systems (FLSs) using KM algorithms frequently implies an additional computational burden that results from the permutation process \cite{wagner2010toward}. This may hinder the deployment of some ML-ELM based on IT2-FELMs (FL) in certain cost-sensitive real world applications.

In the field of Unmanned Aerial Vehicles (UAVs) and image classification, ML-ELM has also been implemented. For instance in \cite{hernandez2020multilayer}, a Multilayer Fuzzy Extreme Learning Machine (ML-FELM) based on the concept of Type-1 FL and H-ELM for active object classification was suggested. Compared to the ML-IT2FELM suggested in \cite{rubio2020multilayer}, the ML-FELM uses type-1 fuzzy sets (FSs) for both feature extraction of the input data, and the classification of the final features. The proposed ML-FELM was implemented in an Unmanned Aerial Vehicle (UAV) for the active classification of different objects at different locations and positions. The classification outcome of the ML-FELM was employed by the UAV as a navigation mechanism to infer its location and distance to each object. This information was exploited by the UAV to grasp and transport each object between two stations in an arena of $2.2 \times 3.3$ metres. The proposed ML-FELM alleviates the problem of reducing the extra computation that results in IT2-FELMs with a KM algorithm by using type-1 fuzzy sets. Based on the experiments presented in \cite{hernandez2020multilayer}, compared to traditional ML-ELM, an ML-FELM provides a higher model performance in the presence of noisy images. However, the computational training of a ML-FELM is still higher than the one required to train a ML-ELM and its performance is lower than a ML-IT2-FELM \cite{rubio2020multilayer}.

Typically, in aerial missions, deep neural structures such as Convolutional Neural Networks (CNNs) have become one of the most recurrent tools to analyse visual imagery \cite{rohan2019convolutional, ma2017hand,  choi2020convolutional}. Deep neural structures such as the CNN usually comprise a feature extractor phase followed by their classification using single-hidden layer neural structures \cite{wu2017introduction, rubio2014interval,tang2015extreme}. For real time applications, efforts to reduce the associated complexity for the training of CNNs and improve their accuracy can be found in literature \cite{ren2015faster, dai2016r, liu2016ssd, redmon2016you}. 

For object detection using UAVs, frequently techniques such as background substraction or Haar cascade classifiers are employed \cite{rohan2019convolutional}. In the application of neural structures such as CNNs, two main categories have been suggested, i.e.: 1) region based detectors and 2) single shot detectors (SSDs) \cite{rohan2019convolutional}. For example in \cite{rohan2019convolutional}, a CNN based on SSD architecture to detect and track predefined target objects was presented. In \cite{rohan2019convolutional}, the proposed neural structure provides an improved accuracy compared to traditional models. As reported in \cite{rohan2019convolutional}, the implementation of SSD detectors usually demand a higher computational load. For that reason, authors implemented the proposed model in a GPU to importantly reduce the associated computational load. In Applications that involve not only object detection but also robotic grasping, various multilayer neural structures have also been suggested. In \cite{lenz2015deep}, authors developed a system for detecting robotic grasps from RGB-D data using a two-cascade deep learning approach. According to the results presented in \cite{lenz2015deep}, the approach requires a small number of features to provide a high balance between model accuracy and low computational load. As detailed in \cite{lenz2015deep}, grasping is inherently a detection problem that involves perception, planning and control. Within this context, a problem that involves autonomous object recognition and their transport is usually treated as a perception problem given a noisy signal with a partial view of the object from a camera. In UAVs, the goal in these type of tasks is to infer locations where a robotic gripper could be placed \cite{lenz2015deep, rohan2019convolutional}. Unlike generic image problems, in autonomous manipulation and grasping of objects, the use of closed loop solutions has been the most popular tool.

This paper extends the work presented in \cite{hernandez2020multilayer} by suggesting a Hybrid Multilayer Extreme Learning Machine (HML-ELM) with an application to active object classification and their transport using and Unmanned UAV. For the recognition of objects, the HML-ELM actively classifies the images collected by the UAV's camera. This information is used by a navigation strategy to guide the UAV to autonomously collect and transport objects with a different shape and orientation between two predefined locations. The proposed HML-ELM is a multilayer neural structure based on the concept of ELM autoencoders (ELM-AEs) and Simplified Interval Type-2 Fuzzy Logic ELM (SIT2-FELM). Compared to the ML-FELM suggested in \cite{hernandez2020multilayer}, the proposed HML-ELM follows a forward hierarchical training approach that involves two main phases not fully based on fuzzy logic theory. First, an unsupervised step is performed to extract a high-feature representation space of the input data. Such representation is achieved by stacking a number of ELM-AEs. In the second phase, classification of the resulting features is carried out by using a SIT2-FELM with an output layer based on the SC type-reduction algorithm \cite{chen2020comprehensive}. On the hand, by using a set of ELM-AEs, the proposed HML-ELM preserves its ability for an improved feature mapping, reducing the time that implies the use of FAEs. On the other hand, by using a simplified version of an IT2-FELM with a SC algorithm, the iterative procedure required in KM algorithms is eliminating reducing importantly the associated time while proving similar model performance.

In the same manner to Deep learning methods based on back-propagation approaches, a HML-ELM extracts features by a multilayer feature representation framework where for feature classification a separate ELM block is applied \cite{tang2015extreme}. In other words, original inputs are mapped into a high feature representation through a process that is decomposed into multiple hidden layers, and the outputs of the previous layer are used as the input of the current one. However, different from greedy layerwise learning in deep learning, a HML-ELM feature representation and classification are two separate blocks that do not require a fine tuning, and thus the associated computational load for their training result much smaller. The proposed HML-ELM follows the learning principles of basic ML-ELM structures where at each hidden layer node parameters are randomly generated and the output weights are calculated analytically. By doing this, at unsupervised stage, encoded output features are randomly projected in a forward manner before final decision in the classification layer is achieved leading to improved generalisation properties on the one hand. On the other hand, similar to IT2 Fuzzy Logic Systems (IT2-FLSs), the final layer in the HML-ELM inherits the ability of IT2-FLSs to better deal with different types of uncertainties usually present in real world applications.  It is known that in a large number of applications, IT2 FLSs outperform their type-1 counterpart and then crisp systems such as ELMs \cite{wagner2010toward}. The final layer of the proposed HML-ELM is a SIT2-FELM which is a blurry version of FELMs where each node in the hidden layer is represented by an Interval Type-2 Fuzzy Set (IT2-FS). Usually the computation of IT2-FLSs implies an extra computational burden due to the type reduction process that is required to process IT2 FSs. The commonly used center of sets type-reducer (COS TR) used in the implementation of IT2-FLSs is the Karnik-Mendel (KM) algorithm and its variants which suffer from the need of a sorting process \cite{chen2020comprehensive}. In this work, the proposed HML-ELM uses a SIT2-FELM with an improved version of the (COS TR) that eliminates the need of sorting and called SC algorithm and thus reduces significantly its training.

To validate the proposed HML-ELM, a number of two different experiments are suggested. First, a number of benchmark image data sets is employed to compared the efficiency of the HML-ELM with respect to other multilayer structures and to standard Convolutional Neural Network (CNN). Secondly, the HML-ELM is integrated in a navigation methodology using UAVs to actively classify and transport objects between two predefined locations. According to experiments, the proposed approach provides a higher trade-off between model accuracy and faster deployment than other deep learning strategies.

The rest of this paper is organised as follows: Section II briefly reviews background theory, and the proposed HML-ELM as well as the navigation strategy to the collection and transport of different objects is presented. Section IV reports experiment results. Finally discussion and conclusions of this research work are described in sections V and VI respectively.

\section{Background Material}
\subsection{Multilayer Extreme Learning Machine (ML-ELM)}

ML-ELM is a hierarchical ELM learning approach to the training of multilayer neural networks that involves two main phases: (1) self-taught feature extraction of input data followed by a (2) supervised classification of these features \cite{tang2015extreme,kasun2013representational}. Self-taught feature extraction is conducted for unsupervised multilayer encoding where a number of $'L'$ ELM-based autoencoders (ELM-AEs) are stacked. This encoding achieves a more compact and meaningful feature representation than traditional ELM. Compared to greedy layerwise training of multilayer neural networks where gradient descent is applied, ML-ELM follows a forward training where fine iterative tuning is not required and the associated computational burden is reduced significantly \cite{kasun2016dimension}. As pointed out in \cite{wang2017encrypted}, compared to traditional ELM, at each ELM-AE, random weights and random biases are chosen to be orthogonal. Such process will make the generalization performance better.  In the unsupervised multilayer encoding, for a given number of distinct samples, each ELM-AE uses the target $\textbf{t}_p = \textbf{x}_p$, where $\textbf{x}_p = [x_1, \ldots, x_N]^T$. The output of each ELM-AE can be calculated as: \cite{wang2017encrypted}:
\begin{equation}
    f(\textbf{x}_p) = g(\textbf{a}\textbf{x} + \textbf{b}) 
\end{equation}
 in which, $\textbf{a}^T\textbf{a} = I$, $\textbf{b}^T\textbf{b}= 1$. The terms $\textbf{a} = [\textbf{a}_1 \ldots, \textbf{a}_M]$ and $\textbf{b} = [\textbf{b}_1 \ldots, \textbf{b}_M]$  are the orthogonal random weights and bias between the input and the hidden layer of each ELM-AE respectively. $g(\cdot)$ is the corresponding activation function of each hidden unit. If sparse or compressed representation of the input data is chosen, the optimal output weights of each ELM-AE are calculated by:
 \begin{equation}
     \pmb{\beta}^{i} =  \left( \frac{\textbf{I}}{C} + (\textbf{H}^{i})^T\textbf{H}^{i} \right)^{-1} (\textbf{H}^{i})^T \textbf{X}^{i},~i=1,\ldots,L
 \end{equation}
where the term $\textbf{h}(\textbf{x}_k) = g(\textbf{x}_k \textbf{A} + \textbf{b}) = [h_1(\textbf{x}_k), \ldots, h_{M_i}(\textbf{x}_k)]$ and $\textbf{X}^i$ is the feature encoding of the $ith$ hidden layer. If it is equal dimension ELM-AE representation of the input data, the optimal output weights are calculated as:
\begin{equation}
    \pmb{\beta}^{i} = (\textbf{H}^{i})^{-1}\textbf{X}^i,~(\pmb{\beta}^i)^T\pmb{\beta}^i = \textbf{I}
\end{equation}
in which, $\textbf{H}^{i} = [\textbf{h}(\textbf{x}_1), \ldots, \textbf{h}(\textbf{x}_N)]^T$ is the output of the $ith$ layer of an ML-ELM. $\textbf{H}^i$ can be calculated as described in Eq. (4)
\begin{equation}
    \textbf{H}^i = g(\textbf{H}^{i-1}\cdot \pmb{\beta}^i)
\end{equation}
And the input data $\textbf{X}^i$ to each hidden layer is computed as:
\begin{equation}
    \textbf{X}^i = f(\pmb{\beta}^i \cdot \textbf{X}^{i-1})
\end{equation}
where $\textbf{X}^0$ is the input data layer, and $\pmb{\beta}^i$ is the output weight of the $ith$ layer. If the number of hidden nodes $'M_i'$ in $ith$ hidden layer  is equal to the number of hidden nodes in $(i-1)th$ layer, $f(\cdot)$ is chosen as linear. Otherwise, $f(\cdot)$ is chosen as nonlinear piecewise, such as sigmoid function $g(\cdot)$. 
\subsubsection{Interval Type-2 Fuzzy Logic System (IT2 FLS)}
Similar to its type-1 counterpart, an IT2 FLS includes a fuzzifier, a rule base, fuzzy inference engine and an output processor as illustrated in Fig. \ref{fig::IT2_FLS}. The output processor consists of a type reducer and a defuzzifier in which a type-1 fuzzy set (T1 FS) and a crisp number is generated respectively. The fuzzifier maps an input vector $\textbf{x}=[x_1, \ldots, x_N]^T \in X_1, \times \ldots \times X_N \equiv \textbf{X}$ into a Type-2 Fuzzy Set (T2 FS) $\tilde{A}_x$ in $\textbf{X}$. An IT2 FLS either of Takagi-Sugeno-Kang (TSK) or Mamdani type is characterised by IF-THEN rules, where their antecedent or consequent sets are usually Interval Type-2 Fuzzy Sets (IT2 FSs) \cite{melin2013review}. For a Multi-Input-Multiple-Output FLS (MIMO FLS) with a Mamdani inference, having $N$ inputs and $\tilde{N}$ outputs $y_i \in Y$, each rule is defined as \cite{rubio2016data}:
\begin{equation}
R^j: \text{IF}~x_{1}~\text{is}~\tilde{A}_{j1} ~\text{AND}~ x_{2}~\text{is}~\tilde{A}_{j2}~\text{AND}~ \ldots 
\text{IF}~x_N~\text{is}~A_{jN}~\text{THEN}~y_i~\text{is}~w_{ij},~j = 1, \ldots, M. 
\end{equation}
where $A_{jk}(k=1,\ldots, N)$ are fuzzy sets of the $kth$ input variable $x_k$ in rule $j$, $N$ and $\tilde{N}$ is the dimension of the input and out vectors respectively. 

Each consequent $w_{ij} =  q_{ij,1} x_1 + \ldots  q_{ij,N} x_N$ if a Takagi-Sugeno-Kang (TSK) inference is used.
In an IT2 FLS, the inference engine combines rules and gives output T2 FSs. 

In a Mamdani (TSK) fuzzy model, the degree to which a given input variable $x_k$ satisfies the quantifier $A_{jk}$ is specified by its membership function $\mu_{A_{jk}}(x_k)$. 

An IT2 FS is characterised by its lower and upper MF  $[\bunderline[0]{ \mu}_{\tilde{A}_{jk} },\oversymb{\mu}_{\tilde{A}_{jk}} ]$ of the Footprint Of Uncertainty (FOU) respectively. The firing strength $\tilde{F}^j$ of each $jth$ fuzzy rule can be obtained by performing fuzzy meet operation with the inputs using an algebraic product operation as follows:
\begin{equation}
\tilde{F}^j = [\bunderline[3]{f}_j(\vec{x}_p),\oversymb{f}_j(\vec{x}_p)] 
\end{equation}
\begin{figurehere}
\begin{center}
\includegraphics[width=10cm,height=4.0cm]{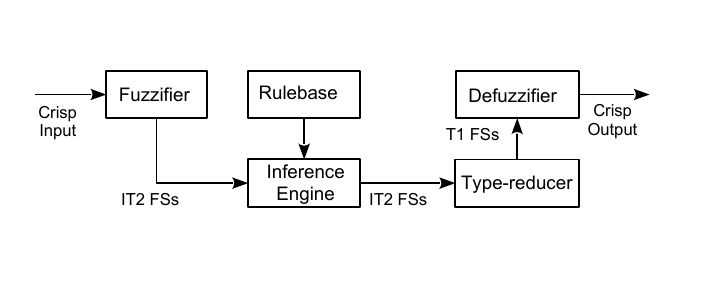}
\caption{\footnotesize Components of an Interval Type 2 Fuzzy Logic System (IT2 FLS).}\label{fig::IT2_FLS}
\end{center}
\end{figurehere}
where terms $[{f}_j(\vec{x}_p),\oversymb{f}_j(\vec{x}_p)]$ are defined by:
\begin{equation}
\bunderline[3]{f}_j(\vec{x}_p) = \prod_{j=1}^M \bunderline[3]{\mu}_{jk},~~\oversymb{f}_j(\vec{x}_p) = \prod_{j=1}^M \oversymb{\mu}_{jk}
\end{equation}
Note that the minimum \textit{t-norm} may also be used in Eq. (4). In a T1 FLS, computing the $ith$ output represents going from a fuzzy set to a crisp number. This process is usually called defuzzification and can be viewed as a mapping of a T1 FS into a number. In an IT2 FLS, computing the $ith$ output, usually two approaches are available: (1) map a T2 FS directly into a number—direct defuzzification, or (2) first map an IT2 FS into a T1 FS (this is called type-reduction) and then map that set into a number (defuzzification)-type-reduction (TR) + defuzzification. In this work, the latter method is implemented, in which TR combines $\tilde{F}^j$ and its corresponding rule consequent $w_{ij}$. In literature, to compute the output of an IT2-FLS many methods can be found \cite{mendel2017uncertain}, the most common used is the center-of-sets type-reducer also adopted in this work:
\begin{equation}
    Y_{COS} = \frac{\sum_{j=1}^{M} Y^j \tilde{F}^j}{\sum_{j=1}^{M} \tilde{F}^j} = [y_l, y_r]_j
\end{equation}
where the type-reduced set $[y_l, y_r]$ for an IT2 FLS with a TSK inference is defined as:
\begin{equation}
  y_l = \frac{\sum_{j=1}^L\oversymb{f}_jw_{ij} + \sum_{j=L+1}^M\bunderline[3]{f}_jw_{ij}}{\sum_{j=1}^L\oversymb{f}_j + \sum_{j=L+1}^M\bunderline[3]{f}_j}
    \label{y_L}
  \end{equation} 
  \begin{equation}
y_r = \frac{\sum_{j=1}^R\bunderline[3]{f}_j w_{ij} + \sum_{j=R+1}^M\oversymb{f}_jw_{ij}}{\sum_{j=1}^R\bunderline[3]{f}_j + \sum_{j=R+1}^M\oversymb{f}_j}
    \label{y_R}
  \end{equation} 
In Eq. (6) and (7), index $j$ is a possible switch point. A common practice to determine such switch points and then the set $[y_l, y_r]_j$ is the use of Karnik-Mendel algorithms (KM) \cite{wu2008enhanced}. To perform defuzzification, the $ith$ output can be obtained as the center of the type reduced $[y_l, y_r]$ as indicated in Eq. (12):
\begin{equation}
    y_i = \frac{y_l + y_r}{2}
\end{equation}
\subsubsection{Interval Type-2 Fuzzy Extreme Learning Machine for Classification (IT2-FELM)}

An IT2-FELM is a learning methodology based on ELM theory for the training of IT2 Fuzzy Logic Systems (IT2 FLSs) functionally equivalent to SLFNs  \cite{deng2013t2fela}. As described in \cite{rubio2020multilayer}, a Multiple-Input-Multiple-Output (MIMO) IT2-FLS usually performs better for the solution of classification problems than its T1 counterpart. 
According to \cite{rubio2020multilayer}, for a MIMO IT2-FLS of TSK (Mamdani) type and given $'P'$ distinct training samples $(\textbf{x}_p, \textbf{t}_p)$, each $y_p^i$ output can be obtained using matrix representation as:
\begin{equation}
y_{p}^i = \frac{1}{2}\left( \textbf{Y}_l^i + \textbf{Y}_r^i \right) \textbf{w}_{i}^T,~i = 1, \ldots, \tilde{N}
\end{equation}
where $y_l^i = \textbf{Y}_l^i \textbf{w}_{i}^T$ and $y_r^i = \textbf{Y}_r^i \textbf{w}_{i}^T$. $\textbf{x}_p = \{x_{p1}, \ldots, x_{pN}\} \in \textbf{R}^{N}$ is an input vector, and $\textbf{t}_p = [t_{p1}, \ldots, t_{p\tilde{N}}]^T \in \textbf{R}^{\tilde{N}}$ the corresponding target. Using an Enhanced version of the KM algorithm (EKM) $\textbf{Y}_l^i$ can be computed as \cite{mendel2004computing}:
\begin{equation}
\textbf{Y}_l^i =  \frac{\oversymb{ \textbf{f}}^T Q_i^T E_{1i}^T E_{1j} Q_i + \bunderline[0]{\textbf{f}}^T Q_i^T E_{2i}^T E_{2i} Q_i}{r_l^T Q_i\oversymb{\textbf{f}} + s_{li}^T Q_i \bunderline[0]{\textbf{f}} }
\end{equation}
$\textbf{w}_{i} = [w_{i1}, \ldots, w_{iM}]^T$ is the set of original rule-ordered consequent weights, and $\textbf{Y}_l^i = (\psi_{li,1}, \ldots, \psi_{li,M})$, and the terms $E_{1i}$, $E_{2i}$, $r_{li}$ and $s_{li}$ are defined as:
\begin{align*}
 E_{1i} &= \left( e_{1i} | e_{2i} | \dots | e_{Li} | \textbf{0}| \dots |\textbf{0} \right)^T~ L_i \times M \\\nonumber
 E_{2i} &= \left(  \textbf{0}| \dots |\textbf{0}| \xi_1^i | \xi_2^i | \dots | \xi_{M - L_i}^i \right)^T~(M-L_i) \times 1 \\\nonumber
 r_{li} &\equiv ( \underbrace{1,1,\dots,1}_{L_i}, 0, \dots, \dots, 0 )^T~M \times 1\\\nonumber
s_{li} &\equiv (0, \dots, \dots, 0 \overbrace{1,1,\dots,1}^{M-L_i} )^T~M \times 1\\\nonumber
\end{align*}
The calculation of the term $\textbf{Y}_r^i$ follows a similar procedure as described in \cite{mendel2004computing}. If a TSK inference is used and based on Eq. (10) and (11), Eq. (9) can be expressed as \cite{rubio2022online}:
\begin{equation}
\begin{split}
y_{p}^i & = \frac{1}{2}\left( y_l^i + y_r^i \right) = \frac{1}{2}\sum_{j=1}^M(\psi_{li,j} + \psi_{ri,j}) w_{ij}\\
 & = \frac{1}{2} \sum_{j=1}^M h_{pj}^i \left( \sum_{k=0}^N x_k q_{ij,k} \right),~ x_{ij,0} = 0, q_{ij,0} = 1;
\end{split}
\end{equation}
Where, $h_{pj}^i = (\psi_{li,j} + \psi_{ri,j})$. Moreover, a compact representation of Eq. (13) can be defined as:
\begin{equation}
y_p^i = \boldsymbol \phi_i \textbf{q}
\end{equation}
where $\textbf{q} = [q_{i1,k},\ldots,q_{i1,k}, \ldots, q_{iM,k},\ldots,q_{iM,k}]^T$ and $\boldsymbol \phi_i$ is
\begin{multline}
   \boldsymbol \phi_p = \frac{1}{2}[(\psi_{li,1}+\psi_{ri,1})x_{p1}, \ldots ,  \psi_{li,1}+\psi_{ri,1})x_{pN}, \ldots \\
   (\psi_{li,M}+\psi_{ri,M})x_{p1}, \ldots ,\psi_{li,M}+\psi_{ri,M})x_{pN}]^T, \in\textbf{R}^{M \times N}
\end{multline}
As indicated in \cite{rubio2020multilayer}, given $P$ training data ($\textbf{x}_p,\textbf{t}$), for each $y_p^i$ a hidden submatrix $\textbf{H}_\textbf{A}$ can be obtained as:
\begin{equation}
 \textbf{H}_{\textbf{A}}(\textbf{x}) = 
  \left( \boldsymbol \phi_{1}~\boldsymbol \phi_{2} ~ \hdots ~\boldsymbol \phi_{P} \right)^T \in \textbf{R}^{P\times (M \times N) }
\end{equation}
According to ELM theory and IT2 Fuzzy Logic, for a multidimensional output $\textbf{T}$, a linear subsystem is required to determine each $ith$ output in the IT2-FELM. At the heart of a TSK FIS, fuzzy modelling can be viewed as a process where the input data space is segmented into fuzzily defined regions which are parameterised and associated with a linear subsystem \cite{rubio2020multilayer}. In other words, a MIMO IT2-FELM can be viewed as a linear combination of a joint block structured pattern that consists of a group of MISO fuzzy models \cite{deng2013t2fela}. Therefore, based on ELM theory, a linear subsystem can be defined for each $ith$ output, where $\textbf{H}_{\textbf{A}}$ can be now called $\textbf{H}_{\textbf{A}}^i$ and calculated as \cite{rubio2020multilayer}:
\begin{equation} 
\textbf{H}_{\textbf{A}}^i(\textbf{x}) \textbf{w}_i = \textbf{t}_p,~\textbf{w}_i \in \textbf{R}^{M \times \tilde{N}}
\label{eq:system}
\end{equation}
Thus, consequent parameters are estimated  with a common block structure over all dimensions of the output variable:
\begin{equation}
    \textbf{H}_{\textbf{B}}^1 \textbf{Q}_1 + \ldots + \textbf{H}_{\textbf{B}}^i \textbf{Q}_i + \ldots + \textbf{H}_{\textbf{B}}^{\tilde{N}} \textbf{Q}_{\tilde{N}} = \textbf{T}
\end{equation}
\[
\textbf{H}_{\textbf{B}}^i \textbf{Q}_i =
\begin{bmatrix}
    h_{11}^i  & \dots & h_{1M}^i \\
    h_{21}^i  & \dots & h_{2M}^i \\
    \vdots  & \dots & \vdots \\
    h_{P1}^i  & \dots & h_{PM}^i
\end{bmatrix}
\begin{bmatrix}
    0  & \dots & w_{1i}   & \ldots  &  0 \\
    0  & \dots & w_{2i}   & \ldots  & 0 \\
    \vdots& \vdots & \vdots & \vdots \\
    0 & \dots & w_{Mi} & \ldots & 0
\end{bmatrix}
\]
\begin{figure*}[t!]
\begin{center}
\includegraphics[width=16cm,height=7.5cm]{
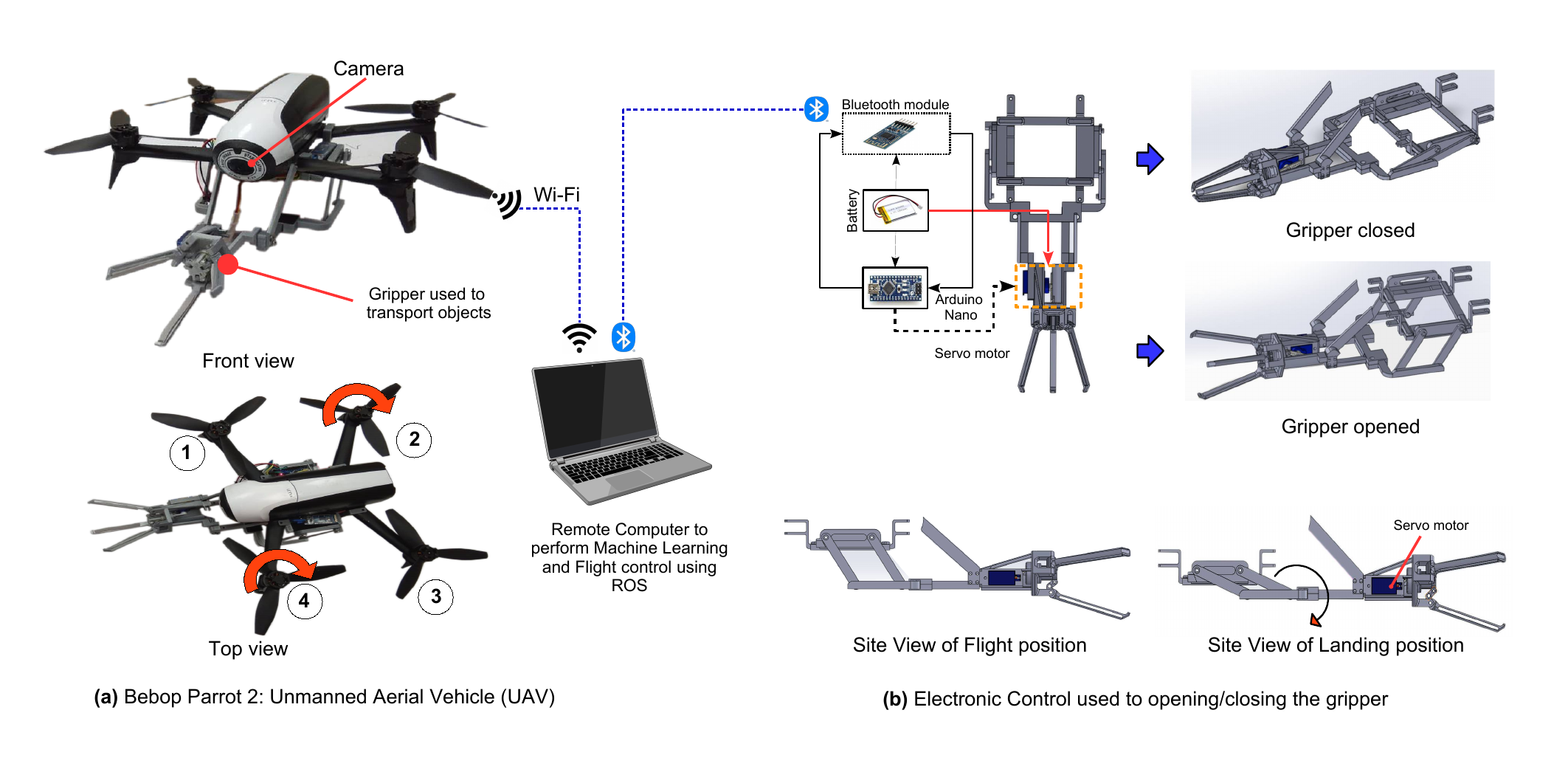}
\vspace{-1mm}
\caption{(a) UAV Bebop2 with a robotic arm and a gripper, (b) Top and site view of the robotic arm and gripper, and (c) Geometrical objects used to perform all experiments.}\label{fig::robotic_platform}
\end{center}
\end{figure*}
where $\textbf{H}_{\textbf{B}}^i \in \textbf{R}^{P \times (M \times N)}$, $\textbf{Q}_i \in \textbf{R}^{(M \times N) \times \tilde{N}}$, and the target $\textbf{T} \in \textbf{R}^{P \times \tilde{N}}$ is a matrix defined as $\textbf{T} = [\textbf{t}_1, \ldots, \textbf{t}_{\tilde{N}}],$ and each $\textbf{t}_i = [t_{1i}, \ldots, t_{Pi}]$. IT2-FELM can be summarised as a three-step learning process \cite{rubio2020multilayer, deng2013t2fela}:
\begin{itemize}
    \item \textbf{Step 1. Random initialisation} of each MF's parameter $m_{jk}$ and $\sigma_{jk}^{1,2}$
    \item \textbf{Step 2. Initialisation of each consequent} $q_{ij,k}$. Calculate the initial value of each consequent $q_{ij,k}$ from the following linear system  ($\tilde{N} \geq 1$):
        \begin{equation}
            \textbf{Q}_A = \textbf{H}_0^{\dagger} \textbf{T}
        \end{equation}
        where $\textbf{H}_0 = [\textbf{h}_1, \ldots, \textbf{h}_P]^T$, $p = 1, \ldots, P$ in which 
\begin{equation}
   \textbf{h}_p = \frac{1}{2}[(y_l^1 + y_r^1) x_{p1}, \ldots ,  (y_l^1+y_r^1) x_{pN}, \ldots
   (y_l^M + y_r^M)x_{p1}, \ldots ,(y_l^M + y_r^M)x_{pN}]^T
\end{equation}
        in which, $\textbf{h}_p \in \textbf{R}^{1 \times (M \times N)}$. To calculate $\textbf{Q}_A$, the value of $y_{l}^i$ and $y_{r}^i$ is obtained as:
        \begin{equation}
            y_l^i = \sum_{j=1}^M \bunderline[3]{f}'_j w_{ij}~, \bunderline[3]{f}'_j = \frac{\bunderline[3]{f}_j}{\sum_{j=1}^M \bunderline[3]{f}}
        \end{equation}
        \begin{equation}
            y_r^i = \sum_{j=1}^M \oversymb{f}'_j w_{ij}~, \oversymb{f}'_j = \frac{\oversymb{f}_j}{\sum_{j=1}^M\oversymb{f}}
        \end{equation}
        Using Eq. (23) and (24), $\textbf{h}_p$ becomes:
\begin{equation}
   \textbf{h}_p = \frac{1}{2}[(\bunderline[3]{f}'_1 + \oversymb{f}'_1) x_{p1}, \ldots ,  (\bunderline[3]{f}'_1 +\oversymb{f}'_1) x_{pN}, \ldots
   (\bunderline[3]{f}'_M + \oversymb{f}'_M)x_{p1}, \ldots ,(\bunderline[3]{f}'_M + \oversymb{f}'_M)x_{pN}]
\end{equation}
        \item \textbf{Step 3. Refinement of each consequent} $q_{ij,k}$. Use the initial matrix $\textbf{Q}_A$ to find the switching points $L_i$ and $R_i$ by applying the EKM to each $ith$ sublinear system. Use this information to build each matrix $\textbf{H}_{\textbf{B}}^i$ and to refine each consequent matrix $\textbf{Q}_i$ by:
        \begin{equation}
            \textbf{Q}_i = (\textbf{H}_{\textbf{B}}^i)^{\dagger} \textbf{t}_i
        \end{equation}
\end{itemize}
\section{Materials and Methods}
\subsection{Aerial Robotic Platform: UAV Bebop Parrot 2 and Gripper}

As illustrated in Fig. \ref{fig::robotic_platform}, the aerial robotic platform used in all experiments consists of two main components, namely: 1) an Unmanned Aerial Vehicle (UAV) called Bebop parrot 2 and 2) a foldable arm with a gripper. As observed in Fig. \ref{fig::robotic_platform}(a), the UAV is equipped with four rotors, the two pairs of propellers (1,3) and (2,4) rotate in opposite directions. Each includes a dedicated brushless direct current motor, a gearbox and a propeller. The Bebop parrot 2 includes in-built PID controllers that allow forward-backward and left-right motions by simultaneously increasing/decreasing the rotor speed. Yaw command is accomplished by accelerating the two clockwise rotors while decelerating the two counter-clockwise rotors. As shown in Fig. \ref{fig::robotic_platform}(c), four different objects with a max weight of 0.035kg are used. For object and environment exploration, the UAV can transmit live video stream from its camera to a remote computer over Wi-Fi connection with a max connection range of 8 mts. The UAV's camera has a resolution of $1920 \times 1080 (1080p)$, where video recording of up to 30 frames (RGB images) per second and max flying time of 25 minutes is used in this work. To control each propeller and estimate the UAV's position and altitude, the bebop autonomy ROS driver was installed in a remote computer to access to IMU data and ultrasound sensor readings. The UAV continuously transmits images  to a remote computer, where active object exploration and classification is performed by the proposed HML-ELM and control methods. To collect each object, a gripping mechanism that consists of a self foldable robotic arm with a gripper was constructed and mounted to the UAV's structure. As illustrated in Fig. \ref{fig::robotic_platform}(b), to improve the gripper's maneuverability for the collection of objects, the design of the robotic arm provides two main configurations, i.e. 1) landing and 2) flight position. The gripper is controlled via Bluetooth with a remote computer. To establish the bluetooth communication between the remote computer and the gripper, a microcontroller Arduino and a Bluetooth module attached to the main structure of the gripper were implemented. To close/open the gripper, control signals from the remote computer are processed by the microcontroller to activate a servo motor with a reducer gearbox which is attached to the gripper structure as shown in Fig. \ref{fig::robotic_platform}(b). To implement all algorithms and control commands to guide the position of the UAV and close/open the gripper as well as the processing of each image, the Robotic Operating System (ROS) middleware in Ubuntu 16.04 was employed.
\begin{figure*}[t!]
\begin{center}
\includegraphics[width=17cm,height=4.4cm]{
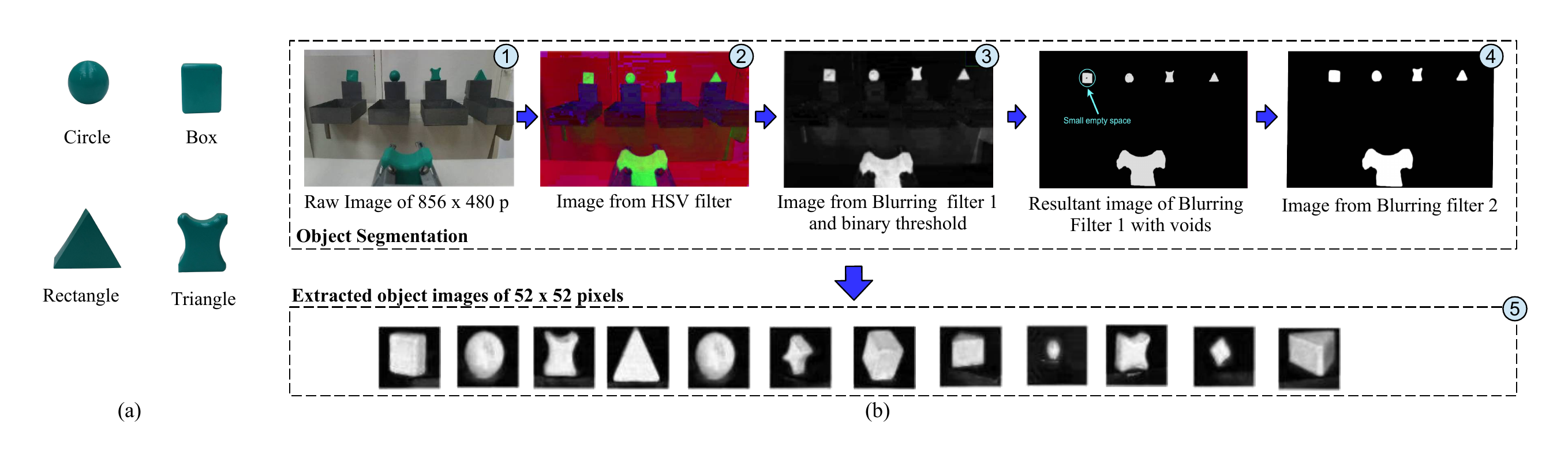}
\caption{\footnotesize (a) Geometric objects used in all experiments and (b) Initial processing of each image transmitted from the unmanned Aerial Vehicle (UAV) to the remote computer.  .}\label{fig::Image_processing_stage}
\end{center}
\end{figure*}

\subsection{Object Data Set Collection and Preprocessing}
To cross-validate the proposed HML-ELM, a final data set of $36386$ images of $52 \times 52$ pixels was collected corresponding to four different objects (called object data for short) as illustrated in Fig. \ref{fig::Image_processing_stage}(a), i.e., box (9114 images), circle (9042 images), irregular shape (9050 images) and triangle (9180 images). To extract the image of each object, a process of image segmentation that consists of six fast steps is performed in each raw image collected by the on-board camera of the UAV. As indicated in Fig. \ref{fig::Image_processing_stage}(b), the process starts by receiving a raw image of $856 \times 480$ from the on-board camera. Consequently, each image is converted from RGB to HSV color space. This process is performed to ensure the system can tolerate to the variations of illumination in the environment used for the experiments. Since the accuracy of color detection affects the results of object classification, a suitable color space for color segmentation is important. Within this context, RGB is not a suitable color space because it is very sensitive to the variations of intensity. Hence, a color space that is insensitive to the strength of illumination is required. A color segmentation based on Hue, Saturation and Value filter (HSV) has been applied. The third step is to apply a blurring filter with a kernel of $3 \times 3$ to extract those features from the Hue layer in a binary image keeping all information related to borders, contour and shape. To eliminate all the information that is not related to each object,  a binary threshold is applied after the blurring filter. As ilustrated in Fig. \ref{fig::Image_processing_stage}, the resultant image of the first blurring filter has an unstable shape produced by the light changes creating small empty spaces/voids inside the object contour. To solve this, a second blurring filter with a kernel size of $7 \times 7$ is applied while the resultant noise in the image is removed out as illustrated in Fig \ref{fig::Image_processing_stage}. Finally, in step 5, the centroid of each object and corresponding coordinates is determined and used to extract an image of each object of $52 \times 52$ pixels. A process for image segmentation is useful to remove out all information that is not relevant to describe each object. Hence, the final image of $52 \times 52$ pixels is converted to a single row vector of $1 \times 2704$ which is fed into the proposed HML-ELM.
\begin{figure*}[t!]
\begin{center}
\includegraphics[width=17cm,height=5.6cm]{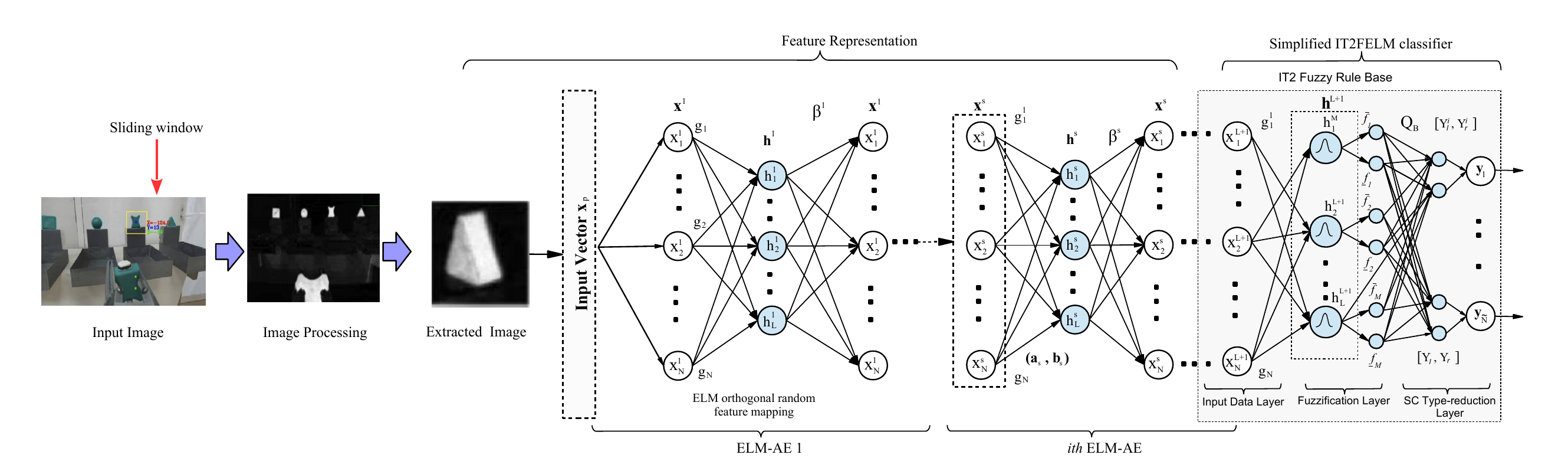}
\vspace{-1mm}
\caption{\footnotesize Proposed Hybrid Multilayer Extreme Learning Machine (HML-ELM) structure.}\label{fig::HML_ELM}
\end{center}

\end{figure*}

\subsection{Proposed Hybrid Multilayer Extreme Learning Machine (HML-ELM)}

The proposed HML-ELM is a hierarchical learning methodology that is based on the original ML-ELM and Simplified Interval Type-2 Fuzzy ELM (SIT2-FELM). As described in Fig. \ref{fig::HML_ELM}, the proposed HML-ELM is a multilayer structure of $'L+1'$ layers that involves two main learning phases. In the initial phase, the first $'L'$ layers are used to perform  unsupervised learning for feature representation of input data. This is achieved by stacking a number of $L$ ELM-based Autoencoders (ELM-AEs), in which each autoencoder can be considered as an independent entity or module. In the second stage, supervised classification of these features is performed by using a Simplified Interval Type-2 Extreme Learning Machine (SIT2-FELM). On the one hand, each ELM-AE can be seen as a special case of traditional ELM where fine tuning is not necessary, and the output weights are determined analytically. On the other hand, supervised classification of extracted features is performed using a SIT2-FELM in which a simplified version of the Center of Set Type Reducer Without Sorting Requirement (COSTRWSR) called SC algorithm is implemented as the output layer. Compared to traditional IT2 Fuzzy Logic Systems with a Karnik-Mendel type reduction layer, the proposed SIT2-FELM eliminates the need of sorting the output weights. By doing this, the associated computational load necessary to train an IT2-FELM is significantly reduced. For a given training data $(\textbf{X}^s,\textbf{T})$, the vector $\textbf{X}^s = [\textbf{x}^s_1, \ldots \textbf{x}^i_P]$ is the input data for each hidden layer such as $s=1, \ldots, L$, $k = 1, \ldots, N$, and $\textbf{T} = [t_1, \ldots, t_P]$ is the target in which $p=1, \ldots, P$. Prior to unsupervised learning, the input raw data should be transformed into an ELM random feature space, in which high data representation of input data can be exploited while maintaining the universal approximation capability of ELM. Similarly to traditional ML-ELM \cite{kasun2013representational}, the output of the first $L-1$ hidden layers of the proposed HML-ELM can be computed by \cite{tang2016extreme}:
\begin{equation}
    \textbf{H}^s = g(\textbf{H}^{s-1}\cdot \pmb{\beta}^s) 
\end{equation}
where $\textbf{H}^{s} = [\textbf{h}^{s}(\textbf{x}^s_1), \ldots, \textbf{h}^{s}(\textbf{x}^s_P)]^T$ is the output matrix of the $ith$ hidden layer w.r.t. input $\textbf{X}^s$,  $g(\cdot)$ denotes the activation function of the hidden layers and $\pmb{\beta}_s$ is the corresponding output weight. As described in Fig. \ref{fig::HML_ELM}, data transformation is achieved by projecting $\textbf{X}^s$ along each output weight $\pmb{\beta}_s$. Mathematically, each ELM-AEs maps input data $\textbf{X}^s$ to a higher level representation, e.g. random input weights and random biases in hidden nodes for additive nodes \cite{kasun2016dimension}. Thus, orthogonal random hidden parameters of linear and non-linear ELM-AE are computed using Eq. (29) \cite{kasun2016dimension}.
\begin{equation}
    \textbf{h}^{s}(\textbf{x}^s_p) = g(\textbf{x}^s_p \textbf{a} + \textbf{b}) = [h_1(\textbf{x}^s_p), \ldots, h_{M_s}(\textbf{x}^s_p)] 
\end{equation}

such as $M_s$ is the number of nodes at the $sth$ hidden layer, $\textbf{a}^T\textbf{a} = \textbf{I}$ and $\textbf{b}^T\textbf{b} = 1$. $\textbf{a}^s = [\textbf{a}_1, \ldots, \textbf{a}_{M_s}]$ are the orthogonal random weight and $\textbf{b} = [b_1, \ldots, b_{M_s}]$ are the orthogonal random bias between the input nodes and hidden nodes \cite{kasun2013representational}. Similarly to an ML-ELM, if sparse or compressed representation of the input data is chosen, the optimal output weights of each ELM-AE are calculated by:
\begin{equation}
    \pmb{\beta}^{s} =  \left( \frac{\textbf{I}}{C} + (\textbf{H}^{s})^T\textbf{H}^{s} \right)^{-1} (\textbf{H}^{s})^T \textbf{X}^{s},~s=1,\ldots,L
\end{equation}
where the term $\textbf{X}^{s}$ is the feature encoding of the $sth$ hidden layer of the proposed HML-ELM. If it is equal dimension ELM-AE representation of the input data, the optimal output weight $\pmb{\beta}^s$ are calculated as:
\begin{equation}
    \pmb{\beta}^{s} = (\textbf{H}^{s})^{-1}\textbf{X}^s,~(\pmb{\beta}^s)^T\pmb{\beta}^s = \textbf{I}
\end{equation}
The final features $\textbf{X}^{L}$ are then computed by \cite{rubio2020multilayer}:
\begin{equation}
    \textbf{X}^{L} = g(\textbf{X}^{L-1} \cdot \pmb{\beta}^{L-1})
\end{equation}
In the second phase, a simplified version of Interval Type-2 Fuzzy Extreme Learning Machine using a SC type reducer (called SIT2-FELM for short) is developed to classify the final features $\textbf{X}^{s+1}$ extracted during the unsupervised learning phase \cite{khanesar2016improving, deng2013t2fela}. The proposed SIT2-FELM is similar to its T1-FELM counterpart, the main difference being that at least one of the FSs in the rule base of the SIT2-FELM is of IT2. Hence, the output of the inference engine of an SIT2-FELM is an IT2 FS, and a type-reducer is required to convert it into a T1 FS before fuzzification. In the deployment of IT2 FLSs, one of the the most important steps in computing the output is the type-reduction (TR) \cite{khanesar2016improving}. A commonly used TR method is the center of sets type reducer (COS TR) which requires the solution of two nonlinear constrained optimisation problems \cite{khanesar2016improving}. A frequent algorithm used to solve this problem is the Karnik Mendel (KM) Algorithm. A TR method based on the KM algorithm and pplied to IT2-FELM is usually intensive in terms of the associate computation \cite{wu2012approaches}. This is mainly due to the iterative nature that results from a sorting process that is performed by KM algorithm for determining the switch points $R$ and $L$. Hence, in certain cost-sensitive real world problems, the application of IT2 FLSs with a KM type-reduction may hinder their deployment.

Many attempts have been reported to improve the efficiency of the KM. Particularly, to reduce the associated computational cost that implies the computation of the sorting process. In \cite{wu2012approaches} and \cite{khanesar2016improving}, a detail review of different TR algorithms is provided categorising them in three main groups:
\begin{itemize}
    \item[1)] Enhanced KM algorithms. These methods improve the original KM algorithms in terms of the associated computational cost \cite{wu2012approaches, duran2008improved, melgarejo2007fast, celemin2013proposal}. However, all these improved versions still require the sorting process to determine the switching points \cite{wu2012approaches}. 
    \item[2)] Close-form type reduction methods. These methods approximate the KM algorithms and eliminate the sorting process by using close forms \cite{wu2012approaches, greenfield2009collapsing,wu2002uncertainty,nie2008towards,coupland2007geometric}.  
    \item[3)] Simplified IT2 FLSs, in which their architecture is simplified by using only a small number of IT2 Fuzzy Sets (FSs, representative) for the most critical input regions while the rest of components employ FSs of T1 are applied \cite{wu2012approaches}. 
\end{itemize}

In this study, the proposed SIT2-FELM methodology incorporates an improved version of the algorithm Center of Sets Type Reducer without Sorting Requirement (COSTRWSR)  \cite{khanesar2016improving} called SC algorithm and suggested in \cite{chen2020comprehensive}. The SIT2-FELM is employed to train a class of IT2FLSs either of Takagi-Sugeno-Kang (TSK) or Mamdani type. As pointed out in \cite{khanesar2016improving}, finding the centroids $y_l$ and $y_r$ can be seen as a process of determining the max and min values of $Y_{COS}$. As it is also indicated in \cite{khanesar2016improving}, Eq. (6) and (7) can be reformulated as:
    \begin{equation}
y_l^i = \frac{ \sum_{j = 1}^{M_{f}} \oversymb{f}_j w_{ij} - \sum_{j=1}^{M_{f}} (1 - z_{lj,i})\Delta u_{ij} w_{ij} }{ \sum_{j = 1}^{M_f} \oversymb{f}_j - \sum_{j=1}^{M_f} (1 - z_{lj,i})\Delta u_{ij}}
\end{equation}
And 
\begin{equation}
y_r^i = \frac{ \sum_{j = 1}^{M_f} \oversymb{f}_j w_{ij} - \sum_{j=1}^{M_f} (1 - z_{rj,i})\Delta u_{ij} w_{ij} }{ \sum_{j = 1}^{M_f} \oversymb{f}_j - \sum_{j=1}^{M_f} (1 - z_{rj,i})\Delta u_{ij}}
\end{equation}
in which, $\Delta u_{ij} = \oversymb{f}_i - \bunderline[3]{f}_i, \forall i \in [1,{M_f}]$ is the difference of the Upper and Lower Membership Functions (UMF, LMF). Where terms $[z_{lj,i},z_{rj,i}]$  can take the values from the interval $[0,1]$, and $w_{ij}$ is the corresponding consequent weight. Moreover, if the values for $[z_{lj,i},z_{rj,i}]$ are taken either equal to $zero$ or $one$, the resulting formula to determine each output $y_p^i$ in an SIT2-FELM can be determined using the SC algorithm as shown in Table 1. Thus, Eq. (24) and (25) are two alternatives to KM algorithms where the need of a sorting process is eliminated. That means, the computing of of terms $L_i$ and $R_i$ do not exist anymore. The SC type reducer is a simplified version  of the Center of Set Type Reducer Without Sorting Requirement Algorithm (COSTRWSR) \cite{khanesar2016improving}. Hence, as detailed in Table I, the SC algorithm finds each $[y_l^n, y_r^n]$ based on a property of derivatives. Compared to COSTRWSR, using the SC algorithm, $[y_l^n, y_r^n]$ can be obtained without adding the extra parameters $[z_{lj,i}, z_{rj,i}]$. However, to provide a matrix representation of the terms $[y_l^i, y_r^i]$, and then implement the concept of ELM, Eq. (31-32) such terms are kept. In this work, each $\bunderline[3]{w}_{ij} = \oversymb{w}_{ij} = w_{ij}$ \cite{khanesar2016improving}. Similar to an IT2-FELM with a KM type reduction, the proposed simplified IT2-FELM contains three main stages:
\begin{itemize}
    \item[1)] \textbf{Generation of antecedents}. Randomly select $m_{jk}$ and $[\sigma_i^1,\sigma_i^2],~i=1, \ldots, M_f$.
    \item[2)] \textbf{Initialisation of each consequent $q_{ij,k}$}. In this stage, the initial value of each consequent $q_{ij,k}$ is determined by approximating the outputs $y_l$ and $y_r$ using the Nie-Tan (NT) method. The NT method is a close-form type-reduction method (direct defuzzification approach) formulated using a vertical-slice representation of IT2 FSs \cite{nie2008towards}. Each vertical slide is an embedded T1 FS that can be easily type reduced. Hence, an NT method computes the centroid (defuzzfication or defuzzified output) of a TSK IT2 FLS by using the following expression \cite{nie2008towards}:
    \begin{equation}
        y_p^i  =   \frac{(\bunderline[3]{f}_j+\oversymb{f}_j)}{\sum_{j=1}^{M_f} \bunderline[3]{f}_j + \sum_{j=1}^{M_f} \oversymb{f}_j} w_{ij}
    \end{equation}
    in which, $ w_{ij} =  q_{ij,0} x_0 + q_{ij,1} x_1 + \ldots  q_{ij,N} x_N $. Let 
   \begin{equation}
   \varphi_{pj} =  \frac{\bunderline[3]{f}_j+\oversymb{f}_j}{\sum_{j=1}^{M_f} \bunderline[3]{f}_j + \sum_{j=1}^{M_f} \oversymb{f}_j}
   \end{equation}
    Use Eq. (27) to compute each input of the hidden matrix $\textbf{H}_0 = [\textbf{h}_1, \ldots, \textbf{h}_P]^T$, where $\textbf{h}_p$ is computed as:
    \begin{equation}
   \textbf{h}_p = [\varphi_{p1} x_{p1}, \ldots ,  \varphi_{p1} x_{pN}, \ldots
   \varphi_{p{M_f}} x_{p{M_f}}, \ldots ,\varphi_{p{M_f}} x_{pN}],~p=1,\ldots,P
   \end{equation}
    Therefore, use $\textbf{H}_0$ in the linear system described in Eq. (18) to calculate the consequent matrix $\textbf{Q}_A$ (initial value for each $q_{ij,k}$). 
    \item[3)] \textbf{Refinement of each consequent}  $q_{ij,k}$. Determine $[z_{lj,i}, z_{rj,i}]$ using the algorithm SC described in Table 1. From Eq. (30) and (31), let 
    \begin{equation}
    \varphi_{pj} = \frac{ \oversymb{f}_j - (1 - z_{lj,i})\Delta u_{ij}  }{ \sum_{j = 1}^{M_f} \oversymb{f}_j - \sum_{j = 1}^{M_f}(1 - z_{lj,i})\Delta u_{ij}} + \frac{ \oversymb{f}_j - (1 - z_{rj,i})\Delta u_{ij} }{ \sum_{j = 1}^{M_f} \oversymb{f}_j - \sum_{j=1}^{M_f} (1 - z_{rj,i})\Delta u_{ij}} 
    \end{equation}
Using Eq. (29), the linear system defined in Eq. (23) can be now reformulated in a simplified expression:
\begin{equation}
    \textbf{Q}_{B} = \textbf{H}_B^{\dagger}\textbf{T}
\end{equation}
in which $\textbf{H}_B = [\textbf{h}_1, \ldots, \textbf{h}_P]^T$, where $\textbf{h}_p$ is defined as described in Eq. (28). Hence, $\textbf{Q}_B$ is the matrix of refined consequent $w_{ij}$ defined as:
\[
\textbf{Q}_B =
\begin{bmatrix}
    w_{11}  & \dots & w_{1{M_f}} \\
    w_{21}  & \dots & w_{2{M_f}} \\
    \vdots  & \dots & \vdots \\
    w_{P1}  & \dots & w_{P{M_f}}
\end{bmatrix}
\]
\end{itemize}
\newpage
\begin{tablehere}
\caption{\footnotesize SC algorithm for computing the end points $y_l^i$ and $y_r^i$ for each output of a SIT2-FELM.}\label{iris_results} 
\centering 
\begin{tabular}{p{2.0cm} p{7cm} p{5cm} }
\hline 
\multicolumn{1}{l}{\scriptsize Step}& \multicolumn{1}{c}{\scriptsize Computing $y_l^i$}  & \multicolumn{1}{c}{\scriptsize Computing $y_r^i$ }\\
 
\hline
\footnotesize \textbf{1}&\multicolumn{2}{c}{\scriptsize If $\bunderline[3]{f}_n = 0, \forall n \in [1,{M_f}]$, then}\\   
\footnotesize &\multicolumn{2}{c}{\scriptsize $y_l^i$ = min($\bunderline[3]{w}_{ij}$), \hspace{3cm}  $y_r^i$ = max($\oversymb{w}_{ij}$) }\\   
\footnotesize &\multicolumn{2}{c}{\scriptsize $\forall j \in [1, {M_f}]$ with $\oversymb{f}_j \neq 0$, Stop}\\
\footnotesize \textbf{2}&\multicolumn{2}{c}{\footnotesize Initialise z$_n = 1$, $\Delta u_{n} = \oversymb{f}_n - \bunderline[3]{f}_n, \forall n \in [1,{M_f}]$  }\\ 
\footnotesize \textbf{3}&\multicolumn{2}{c}{\footnotesize Calculate: }\\
&\multicolumn{2}{c}{\footnotesize $\Bigg\{ \delta_1 = \displaystyle \sum_{n = 1}^{M_f}  \oversymb{f}_j, \delta_{2} = \sum_{n = 1}^{M_f} \oversymb{f}_j w_{in}$\Bigg\} }\\
\footnotesize \textbf{4}&\multicolumn{2}{c}{\footnotesize $flag = 0$ }\\
\footnotesize \textbf{5}&\multicolumn{2}{c}{\footnotesize For $j = 1$ to ${M_f}$, repeat the following operations of this step }\\
\footnotesize &\multicolumn{2}{c}{\footnotesize $A_j = w_{ij} \delta_1 - \delta_2$ }\\
&\footnotesize If $A_j < 0$ & \multicolumn{1}{c}{\footnotesize If $A_j >0$ }\\ &\multicolumn{2}{c}{\scriptsize z$'_{j} =1$, else z$'_{j} =0$ } \\
&\multicolumn{2}{c}{\footnotesize If z$'_{j} \neq z_{j}$ then} \\
&\multicolumn{2}{c}{\footnotesize If z$_j = 1,$ $
    \begin{cases}
     flag = 1,~\delta_{1} = \delta_{1}  + \Delta u_{j} \\
 \text{z}_{j} = \text{z}'_j,~~~\delta_{2}=\delta_{2} + u_j \Delta u_{j}
    \end{cases}$ }  \\
&\multicolumn{2}{c}{\footnotesize Else $
    \begin{cases}
     flag = 1,~\delta_{1} = \delta_{1}  - \Delta u_{j} \\
 \text{z}_{j} = \text{z}'_j,~\delta_{2}=\delta_{2} - u_j \Delta u_{j}
    \end{cases}$ }  
\\
\footnotesize \textbf{6}&\multicolumn{2}{c}{\footnotesize If $flag \neq 0$ \footnotesize go to step \textbf{4}; else}\\
&\multicolumn{1}{c}{\footnotesize $y_l^i = \delta_2/\delta_1$; $z_{lj,i} = z_j$}& \multicolumn{1}{c}{\footnotesize $y_r^i = \delta_2/\delta_1$; $z_{rj,i} = z_j$}\\
\hline
		\end{tabular}
		\centering 
		\label{table:evolution_results} 
\end{tablehere}
\vspace{6mm}


\section{Experiments and Results}
For the validation of the proposed HML-ELM, two different types of experiments are implemented in this section. First, three popular benchmark data sets usually employed to evaluate the performance of deep learning methods are used to test the performance of the proposed HML-ELM. As part of this first experiment, the object data is also used to compare the performance of the proposed HML-ELM with other existing techniques.  The second experiment involves a practical application of the proposed HML-ELM to active object recognition with Unmanned Aerial Vehicles (UAVs). All classification simulations related to experiment one were carried out in a macOS laptop with a core i7 and a processor of 2.7GHz using MATLAB 2016a. For the second experiment, a Lenovo computer with a core i7 processor was used as a remote computer to communicate the UAV and all machine learning models implemented in this study. The middleware Robot Operating System (ROS) was installed in the remote computer to run all related machine learning algorithms where MATLAB, Python and C++ were the main coding languages.  

\subsection{Experiment 1: Benchmark Image Data Sets}
In this section, the performance of the proposed HML-ELM is evaluated with respect to other existing methodologies to the solution of three popular benchmark data sets and the object data set created in this study. The first three data sets include a) MINIST \cite{lecun1998mnist}, b) CIFAR-10 \cite{recht2018cifar} and c) MNIST fashion \cite{xiao2017fashion}. The MNIST and fashion-MNIST data sets consist of a total of $70,000$ handwritten digits and clothes of $28 \times 28$ pixels images  respectively ($60,000$ for training and $10,000$ and testing). Both data sets contain greyscale images, hence the input layer to each model is of size $N = 784$ associated with a label for $10$ classes. Similarly, the CIFAR-10 data set consists of $60,000$ records for ten classes, where each image is in RGB format. For all images, each of the three channels are treated as adding additional input pixels \cite{tissera2016deep}, and hence $N = 32 \times 32 \times 3 = 3072$. For cross-validation purposes, the CIFAR-10 data set was split into two sets, $50,000$ and $10,000$ images are for training and  testing correspondingly. The fourth set is the object data fully described in section 3. The object data is a result of 36386 images of $856 \times 856$ pixels, where $70\%$ and $30\%$ was used for training and testing respectively. The accuracy metric is used to determine any model's prediction performance. This metric which is defined in Eq. (39) is the ratio of the number of correctly classified instances to the number of all instances.
\begin{equation}
    Accuracy = (TP + TN)/(TP + TN + FP + FN)
\end{equation}

where TP and FP is true positive and false positive predicted value respectively, while TN and FN is true negative and false negative predicted value correspondingly.

\begin{tablehere}
\caption{\scriptsize Average model accuracy of ten trials for object classification.}\label{iris_results} 
\centering 
\begin{tabular}{p{2cm} |p{1cm}|p{1cm} |p{1cm}|p{1cm} |p{1cm}| p{1cm} | p{4cm} }
\hline 
 \scriptsize Models &\multicolumn{3}{c}{\scriptsize Training ($\%$)}  & \multicolumn{3}{c}{\scriptsize Testing ($\%$)} & \scriptsize Net. structure\\
 
\hline
& \multicolumn{2}{c}{\scriptsize Mean}&\scriptsize Time (s)&\multicolumn{3}{c}{\scriptsize Mean}\\   

\hline 
\multicolumn{8}{c}{\scriptsize  \textbf{MNIST} } \\
\hline
 \scriptsize  CNN& \multicolumn{2}{c}{\scriptsize 99.85} & \multicolumn{1}{c}{\scriptsize 2325} &\multicolumn{3}{c}{\scriptsize 99.66} & \multicolumn{1}{c}{\scriptsize 3 hidden layers} \\
 \scriptsize  ML-ELM& \multicolumn{2}{c}{\scriptsize 99.41} & \multicolumn{1}{c}{\scriptsize 446.31} &\multicolumn{3}{c}{\scriptsize 99.12}& \multicolumn{1}{c}{\scriptsize [784, 700 $\times$ 2, 15000, 10] } \\
 \scriptsize  ELM & \multicolumn{2}{c}{\scriptsize 98.85} & \multicolumn{1}{c}{\scriptsize 602.1} &\multicolumn{3}{c}{\scriptsize 98.37} & \multicolumn{1}{c}{\scriptsize [780] }\\
 \scriptsize  HML-ELM& \multicolumn{2}{c}{\scriptsize 99.77} & \multicolumn{1}{c}{\scriptsize 1430.3} &\multicolumn{3}{c}{\scriptsize 99.62} & \multicolumn{1}{c}{\scriptsize [784, 700 $\times$ 2, 9200, 10] }\\
  \scriptsize  ML-IT2FELM& \multicolumn{2}{c}{\scriptsize 99.70} & \multicolumn{1}{c}{\scriptsize 3119.0} &\multicolumn{3}{c}{\scriptsize 99.59}&\multicolumn{1}{c}{\scriptsize [784, 700 $\times$ 2, 9200, 10] }\\
  \scriptsize  ML-FELM& \multicolumn{2}{c}{\scriptsize 99.36} & \multicolumn{1}{c}{\scriptsize 1129.3} &\multicolumn{3}{c}{\scriptsize 99.10}&\multicolumn{1}{c}{\scriptsize [784, 700 $\times$ 2, 9200, 10] }\\
\hline
\multicolumn{8}{c}{\scriptsize  \textbf{CIFAR-10} } \\
\hline
 \scriptsize  CNN& \multicolumn{2}{c}{\scriptsize 64.19} & \multicolumn{1}{c}{\scriptsize 1704.3} &\multicolumn{3}{c}{\scriptsize 63.29} & \multicolumn{1}{c}{\scriptsize 3 hidden layers } \\
 \scriptsize  ML-ELM& \multicolumn{2}{c}{\scriptsize 57.52} & \multicolumn{1}{c}{\scriptsize 475.81} &\multicolumn{3}{c}{\scriptsize 54.92}& \multicolumn{1}{c}{\scriptsize [3072,970 $\times$ 2, 9200,10] } \\
 \scriptsize  ELM &  \multicolumn{2}{c}{\scriptsize 55.22 } & \multicolumn{1}{c}{\scriptsize 7.3} &\multicolumn{3}{c}{\scriptsize 53.28 } &  \multicolumn{1}{c}{\scriptsize [3550] }\\
 \scriptsize  HML-ELM& \multicolumn{2}{c}{\scriptsize 57.96} & \multicolumn{1}{c}{\scriptsize 905.1} &\multicolumn{3}{c}{\scriptsize 55.48} & \multicolumn{1}{c}{\scriptsize [3072,970 $\times$ 2, 9300,10]}\\
  \scriptsize  ML-IT2FELM& \multicolumn{2}{c}{\scriptsize 57.70} & \multicolumn{1}{c}{\scriptsize 1701.9} &\multicolumn{3}{c}{\scriptsize 55.90} & \multicolumn{1}{c}{\scriptsize [3072,970 $\times$ 2, 9300,10]}\\
  \scriptsize  ML-FELM& \multicolumn{2}{c}{\scriptsize 57.71} & \multicolumn{1}{c}{\scriptsize 604.4} &\multicolumn{3}{c}{\scriptsize 55.09}&\multicolumn{1}{c}{\scriptsize [3072, 700 $\times$ 2, 9300, 10] }\\
\hline
\multicolumn{8}{c}{\scriptsize  \textbf{Fashion-MNIST} } \\
\hline
 \scriptsize  CNN& \multicolumn{2}{c}{\scriptsize 91.78} & \multicolumn{1}{c}{\scriptsize 2922.1} &\multicolumn{3}{c}{\scriptsize 91.49} & \multicolumn{1}{c}{\scriptsize 3 hidden layers } \\
 \scriptsize  ML-ELM& \multicolumn{2}{c}{\scriptsize 92.01} & \multicolumn{1}{c}{\scriptsize 503.1} &\multicolumn{3}{c}{\scriptsize 91.06}& \multicolumn{1}{c}{\scriptsize [784, 950$\times 2$,8900, 10]} \\
 \scriptsize  ELM &  \multicolumn{2}{c}{\scriptsize 86.19} & \multicolumn{1}{c}{\scriptsize 20.19} &\multicolumn{3}{c}{\scriptsize 86.04} & \multicolumn{1}{c}{\scriptsize [700] }\\
 \scriptsize  HML-ELM& \multicolumn{2}{c}{\scriptsize 91.98} & \multicolumn{1}{c}{\scriptsize 1117.9} &\multicolumn{3}{c}{\scriptsize 91.36} & \multicolumn{1}{c}{\scriptsize [784, 950$\times 2$,8900, 10]} \\
  \scriptsize  ML-IT2FELM& \multicolumn{2}{c}{\scriptsize 92.67} & \multicolumn{1}{c}{\scriptsize 4220.3} &\multicolumn{3}{c}{\scriptsize 91.51}& \multicolumn{1}{c}{\scriptsize [784, 950$\times 2$,8900, 10]} \\
  \scriptsize  ML-FELM& \multicolumn{2}{c}{\scriptsize 91.44} & \multicolumn{1}{c}{\scriptsize 533.1} &\multicolumn{3}{c}{\scriptsize 90.76}&\multicolumn{1}{c}{\scriptsize [784, 950 $\times$ 2, 8900, 10] }\\
  \hline
\multicolumn{8}{c}{\scriptsize  \textbf{Object data set} } \\
\hline
 \scriptsize  CNN& \multicolumn{2}{c}{\scriptsize 99.97} & \multicolumn{1}{c}{\scriptsize 5391.1} &\multicolumn{3}{c}{\scriptsize 99.14} & \multicolumn{1}{c}{\scriptsize 3 hidden layers} \\
 \scriptsize  ML-ELM& \multicolumn{2}{c}{\scriptsize 95.32} & \multicolumn{1}{c}{\scriptsize 340.2} &\multicolumn{3}{c}{\scriptsize 94.16}& \multicolumn{1}{c}{\scriptsize [1740, 1600 $\times$ 2, 3000, 4] } \\
 \scriptsize  ELM & \multicolumn{2}{c}{\scriptsize 91.33} & \multicolumn{1}{c}{\scriptsize 29.4} &\multicolumn{3}{c}{\scriptsize 88.80} & \multicolumn{1}{c}{\scriptsize [1600] }\\
 \scriptsize  HML-ELM& \multicolumn{2}{c}{\scriptsize 98.46} & \multicolumn{1}{c}{\scriptsize 980.3} &\multicolumn{3}{c}{\scriptsize 97.60} & \multicolumn{1}{c}{\scriptsize [1740, 1600 $\times$ 2, 3000, 4] }\\
  \scriptsize  ML-IT2FELM& \multicolumn{2}{c}{\scriptsize 98.29} & \multicolumn{1}{c}{\scriptsize 3001.0} &\multicolumn{3}{c}{\scriptsize 98.01}&\multicolumn{1}{c}{\scriptsize [1740, 1600 $\times$ 2, 3000, 4] }\\
  \scriptsize  ML-FELM& \multicolumn{2}{c}{\scriptsize 97.47} & \multicolumn{1}{c}{\scriptsize 629.2} &\multicolumn{3}{c}{\scriptsize 97.30}&\multicolumn{1}{c}{\scriptsize [1740, 1600 $\times$ 2, 3000, 4] }\\
\hline
		
		\end{tabular}
		\centering 
		\label{nfm_results} 
\end{tablehere}
\vspace{5mm}

As detailed in Table 2, on the one hand, a comparison of the performance between the HML-ELM  and methodologies such as a CNN \cite{huang2011extreme,kasun2013representational}, ELM \cite{huang2011extreme} and ML-ELM \cite{tang2015extreme} is provided. On the other hand, to compare the HML-ELM with other fuzzy logic ELM approaches, in Table 2, the performance provided by an ML-IT2FELM \cite{rubio2020multilayer} and ML-FELM \cite{hernandez2020multilayer} is also presented. An ML-IT2FELM is a three-hidden layer network with two IT2 fuzzy autoencoders (FAEs) and one IT2-FELM with a KM algorithm as the final classification layer. A ML-FELM is based on type-1 FL which consists of two fuzzy autoencoders (FAEs) of type-1 and a FELM classifier. To evaluate the performance of each classifier, the average accuracy of ten random experiments is presented in Table 2. The accuracy of each model presented in Table 2 is the average value computed from the resultant confusion matrix. 

Similar to the HML-ELM, the training of the ML-IT2FELM is based on ELM. In this work, the HML-ELM with two ELM-AEs and an IT2-FELM with a SC type-reducer was implemented for all experiments.

The arrangement $C_s=[10^{-1}, 1^{4}, 1^{200}]$, $C_s=[10^{-1}, 1^{4}, 1^{20}]$ and $C_s=[10^{-2}, 1^{4}, 1^{90}],~ \left( s = 1, \ldots, L\right)$ was used for the HML-ELM to classify the MNIST, CIFAR-10 and fashion-MNIST data sets respectively. In Table 2, last column describes the configuration structure used by each model. Based on the experiments presented in Table 2, it can be observed that the faster method is an ELM which in turns provides the poorest performance. In contrast, the CNN provides the highest performance with the largest computational training load. 

 A ML-ELM can be treated as a crisp version of an ML-IT2FELM which in terms of testing accuracy produces the highest model accuracy among all ML-ELM structures. However, this can be compensated by the HML-ELM's performance in terms of a higher trade-off between model accuracy, lower computational training load and model simplicity. As shown in Table 2,  the number of parameters of the HML-ELM is smaller compared to an ML-ELM and much lower with respect to a CNN while obtaining similar a performance. 

 In order to compare the accuracy of the proposed HML-FELM for the classification of the object data set, the experimental setup for the Convolutional Neural Network (CNN) with a structure that consists of  convolution($48 \times 48 \times 32$)-pooling($24 \times 24 \times 32$)-convolution($22 \times 22 \times 32$)-convolution($20 \times 20 \times 64$)-pooling($10 \times 10 \times 64$)-classifier($64000-500-4$). It was found that a value for $C_s=[1^{3},1^{7},1^{49}]$ and $C_s=[1^{2},1^{14},1^{40}]$ provides the highest balance between training and testing accuracy for the HML-FELM and ML-ELM respectively. These regularization parameters were also used for the ML-FELM and ML-IT2FELM.
 
 \begin{figure*}[t!]
\begin{center}
\includegraphics[width=5.3cm,height=4.5cm]{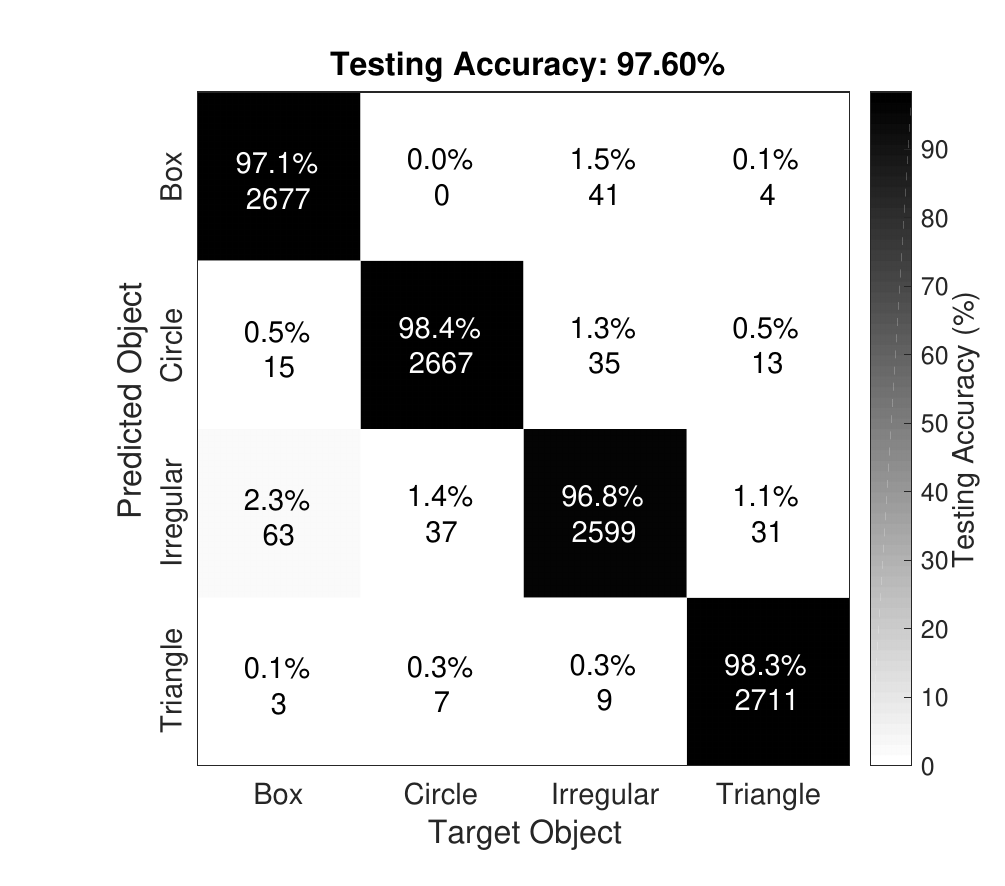}
\vspace{-1mm}
\caption{Model accuracy of a HML-ELM for (a) testing accuracy using $30\%$ out of the original object data set. }\label{fig::classification_results}
\end{center}
\end{figure*}
\vspace{5mm}
 According to our results, the highest model accuracy for the classification of the object data set is achieved by the proposed CNN. The second best performance is provided by the proposed HML-ELM. However, the highest balance between model accuracy and mocel simplicity is provided by the HML-ELM. As presented in Table 2, the computational load associated to the training of a CNN is almost six times the time for training a HML-ELM, which in turn needs many less parameters to produce a similar performance. To illustrate the average testing accuracy achieved by the HML-ELM to the classification of each object, in Fig. \ref{fig::classification_results}, the corresponding average confusion matrix of ten experiments is illustrated. From Fig. \ref{fig::classification_results}(a), it can be observed that the lowest and highest accuracy is produced for the recognition of irregular figures and circle respectively.
\begin{figure*}[!t]
\begin{center}
\includegraphics[width=17cm,height=7cm]{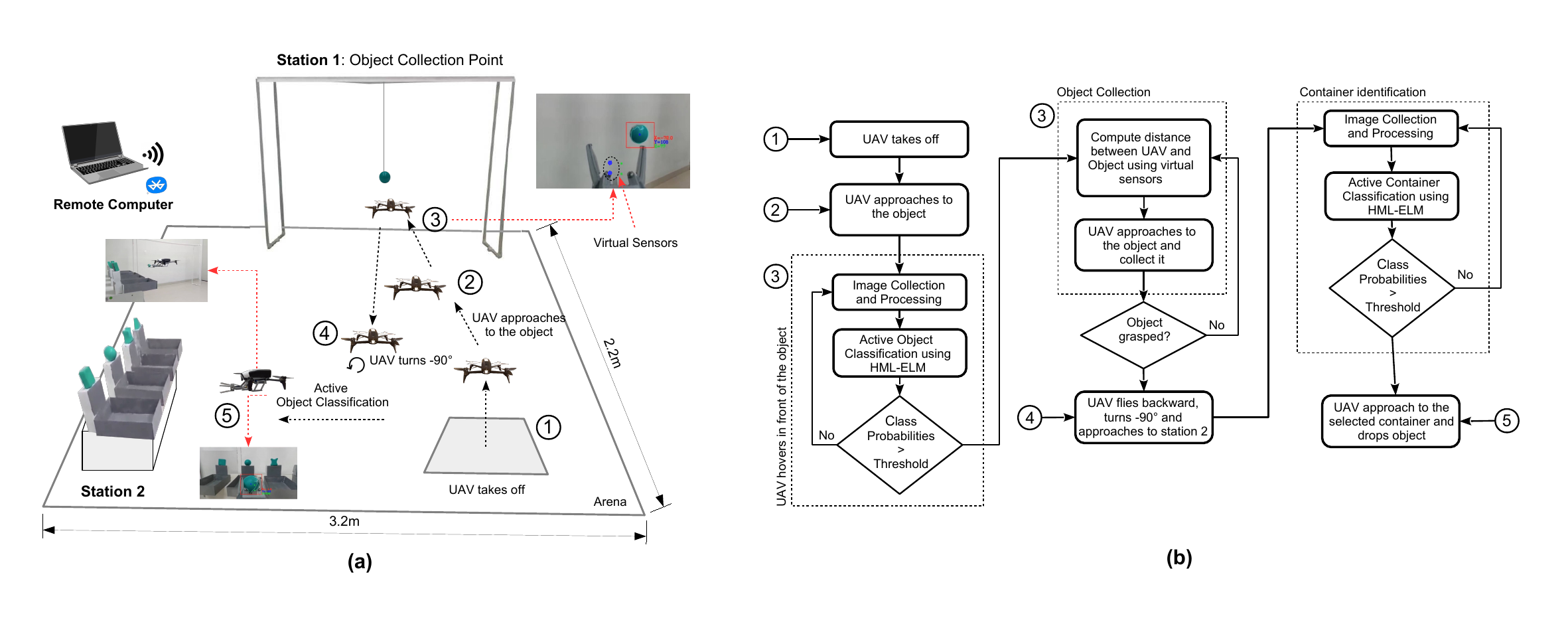}
\caption{\footnotesize (a) Navigation task to be completed by an UAV to the collection and transport of four different objects and (b) proposed navigation strategy to actively classify and transport four different objects .}\label{fig::drone_task}
\end{center}
\end{figure*}
\subsection{Experiment 2: Active Classification and Transport of Objects using an Unmanned Aerial Vehicle (UAV)}

In this experiment, the proposed HML-ELM is used as an active classification mechanism under real conditions. The HML-ELM is implemented in a UAV as the main classification engine to autonomously complete a free-obstacle mission that consists of identifying, collecting and transporting a number of four different objects between two predefined positions (stations 1 and 2). All experiments are carried out in an arena of $2.3mts \times 3.2mts$ as shown in Fig. \ref{fig::drone_task}(a). As described in Fig. \ref{fig::drone_task}, the navigation strategy to actively classify and transport objects follows five main steps. First, the 1) UAV takes off and determines the location of station 1, where an object has been placed and held by a plastic structure. In the second step, 2) the UAV approaches to the station and hovers in front of the object. In step 3) three actions are performed:
\begin{itemize}
    \item[I)] \textbf{Image transmission}. The on-board camera of the UAV continuously collects images and transmit them to a remote computer at a rate of 10Hz using wifi connection.
    \item[II)] \textbf{Image preprocessing and classification}. Before each image is fed into the proposed HML-ELM, a filtering process that converts each image from RGB to HSV color space is performed as detailed in section 3. This process eliminates all information that is not related to each object. From each resultant image, the centroid of each object and its coordinates are determined in order to extract a final image of $52 \times 52$ pixels.  Once each image is fed and classified by the HML-ELM, The probability of each classification is obtained as the number of times the output of HML-FELM ($Y_{HML-FELM}$) is an object $o_m$ class divided by the total number of images used to identify an object .
\begin{equation}
P(c|o_j) = \frac{number~of~times~Y_{HML-FELM} ~is~o_j}{number~of~collected~images};~j=1,\ldots,\tilde{N}
\end{equation}
where $j = 1, \ldots, C$, such as $C$ is the number of object classes, and c the current class. Thus, the classification of the current object is obtained with the current maximum a posterior (MAP) estimate as:
\begin{equation}
\hat{c} = \argmax_{c} P(c|o_i)
\end{equation}
If the value of $\hat{c}$ is higher than a predefined threshold $t_c$, then this information is used to guide the drone to identify the container that corresponds to the classified object. On the contrary, this process is repeated until the value of $\hat{c}$ is satisfied.
    \item[III)] \textbf{Object grasping}. After object classification, to determine the distance between the UAV and each object, two virtual sensors have been setup as visual references as detailed in Fig. \ref{fig::A1}(a). The two visual sensors are fixed marks artificially created in each image with predefined coordinates ($x_1, y_1$) and used by the UAV to control its position with respect to the object. Based on this information, a Proportional Derivative controller is suggested to control the UAV's position. According to our experiments, it was found that using three different values for constants K and P based on the size of each object provides the most suitable control. As described in Fig. \ref{fig::A1}(b) and (c), according to the size of the object three reference marks are established, i.e. position 1, 2 and 3.  This allows the UAV to response faster when is closer to each object. To determine the parameters of the proposed PD control, the value for gains K and P are calculated experimentally. Thus, the control law employed by the UAV is defined by: 
\begin{equation}
\dot{X}=(X_{ref}-X_{act}) \times KX_{L}
\end{equation}
in which, $x$ is the current position and ($X_{ref}$) is the distance between the object and the UAV's position where ($X_{act}$) is the gain $KX_L$ for each position as shown in Fig. \ref{fig::A1}(b) or (c). A similar model for the axis Y and Z were implemented. 
\end{itemize}
In step 4) once the object has been identified and collected, the UAV flies towards the take-off zone where it turns left $90^o$ depicted in Fig. \ref{fig::drone_task}. From this location, the UAV flies to location four where the distance between the UAV and the containers is computed once more by the visual sensors. In order to determine which is the correct target container, in step 5) four objects with the same shape that those objects that are recognised and transported by the UAV are place at the top of each container (See Fig. \ref{fig::drone_task}). In other words, each object is used as a label to each container. Hence, the output of the proposed HML-ELM is actively used to control not only the current position of the UAV, but also the gripper opening and closing. In a like-manner to the third step, a number of predefined images are collected from the label of each container, and then preprocessed to feed the HML-FELM. A new value of $\hat{c}$ is then computed and used to guide the UAV to the goal destination, where the robotic gripper drops the selected object. As detailed in Fig. \ref{fig::drone_task}(b), the proposed navigation methodology involves three basic behaviours, i.e. 1) active image classification, 2) proximity between an object and the UAV using the visual sensors, and 3) (grasp) collection, and delivery of a payload. To illustrate the ROS configuration described by the proposed methodology and related controllers, in Fig. \ref{fig::ROS_configuration} the node architecture for autonomous navigation is illustrated.


\begin{figurehere}
\begin{center}
\includegraphics[width=16cm,height=4.5cm]{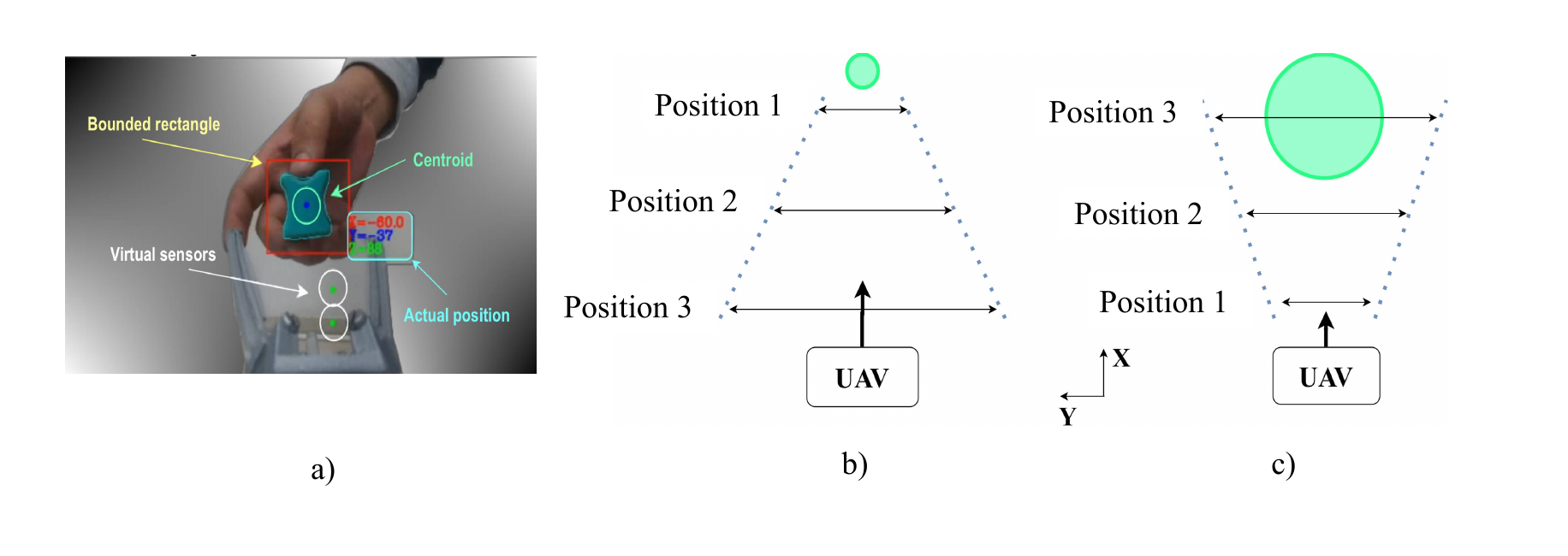}
\caption{\footnotesize a) Virtual sensors created on each image to determine the distance between the UAV and closest object, and b) reference distances used estimate parameters of PD controllers.}\label{fig::A1}
\end{center}
\end{figurehere}
\begin{figurehere}
 \begin{center}
 \includegraphics[width=13cm,height=5cm]{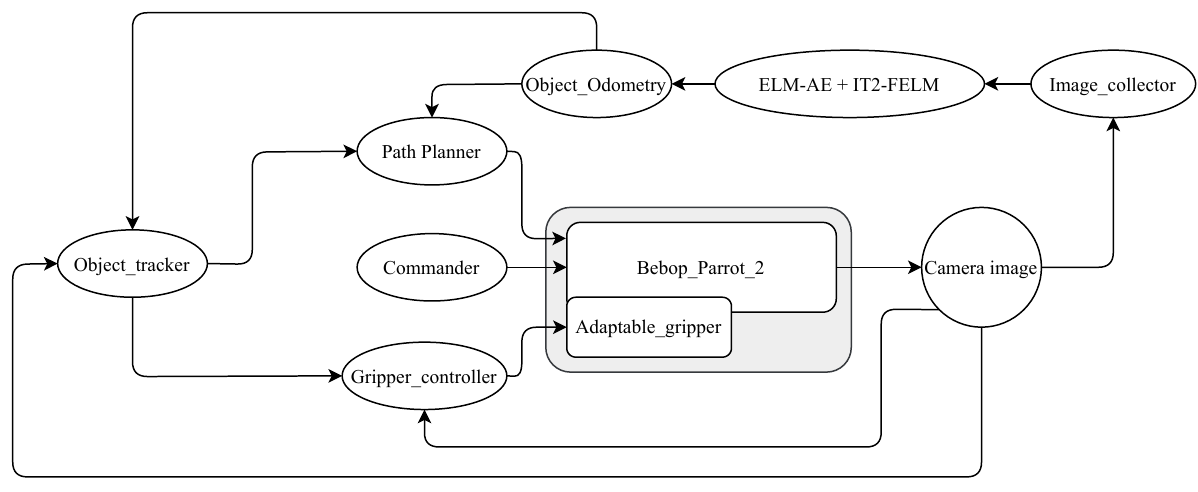}
 \caption{\footnotesize ROS node configuration used for autonomous navigation .}\label{fig::ROS_configuration}
 \end{center}
 \end{figurehere}

\subsection{Results for Active Object Classification}
To validate the proposed HML-ELM for real time object classification, in this section the average of a number of five successful experiments is reported. As described in previous section, each experiment consists of the recognition, collection and transport of an object. In Table 3, the average object classification provided by different methodologies and integrated in the robotic platform is described using a classification threshold of $t_c(\hat{c})= 0.82$. Column Image classification time refers to the average time necessary to recognise an object. Column active classification is the average classification of 120 images collected by the UAV at stations one and two at different positions and orientations in each experiment.  

\begin{figure*}[t!]
\begin{center}
\includegraphics[width=5.3cm,height=4.5cm]{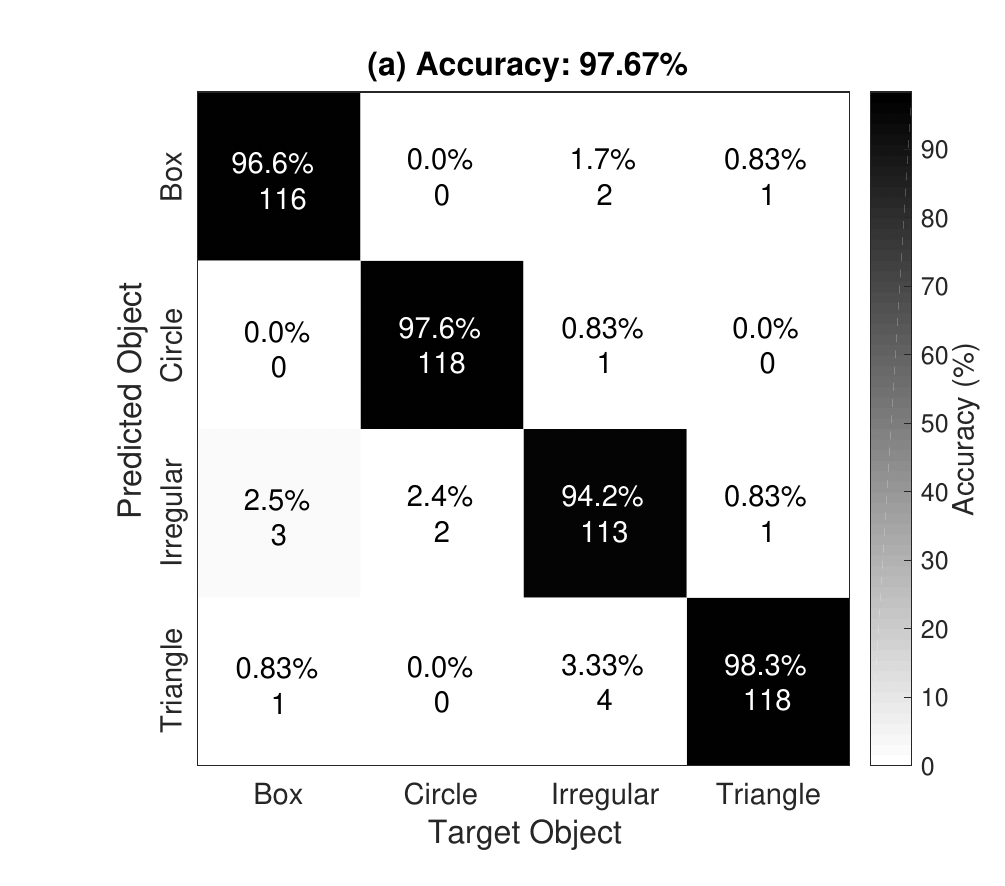}
\includegraphics[width=5.3cm,height=4.6cm]{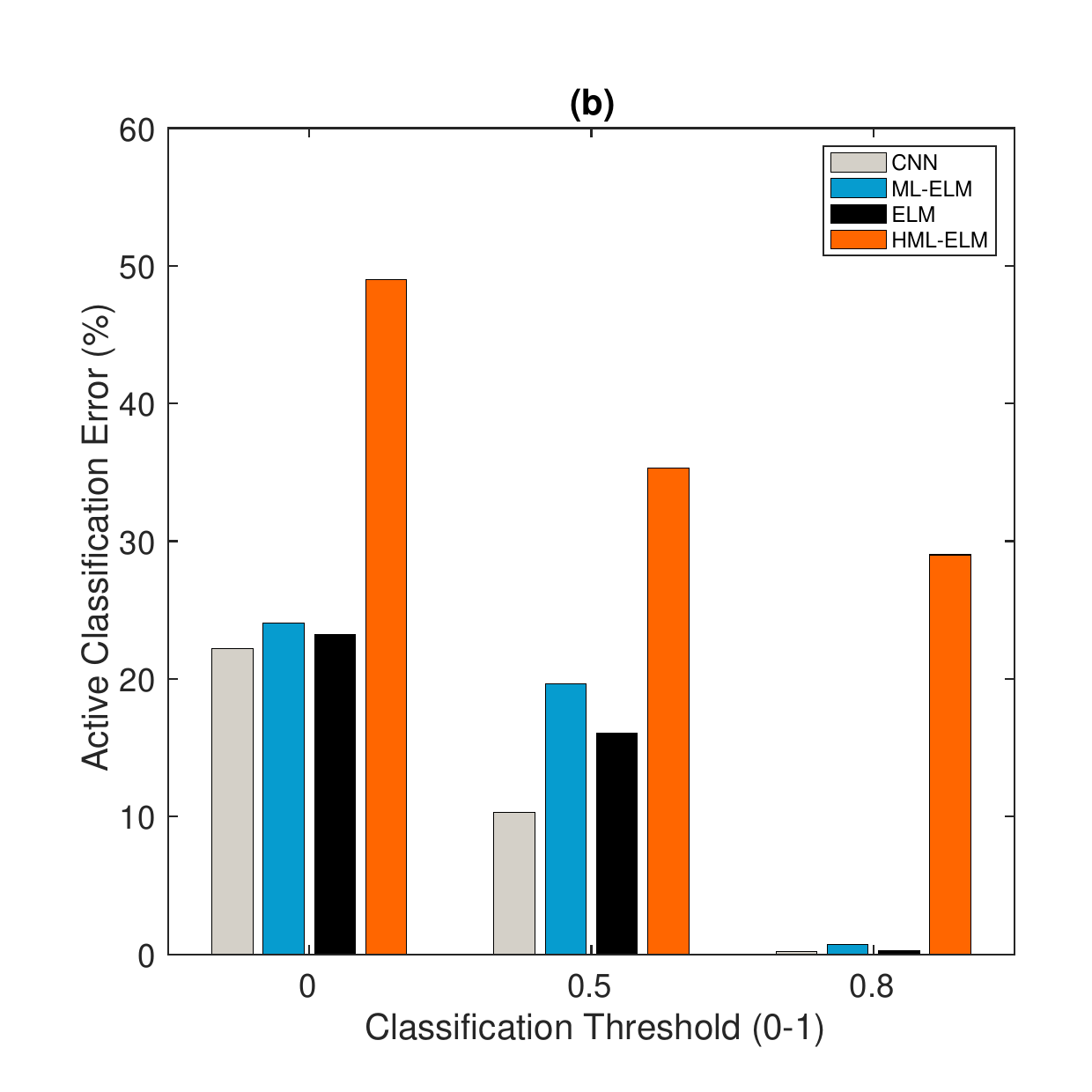}
\includegraphics[width=5.3cm,height=4.6cm]{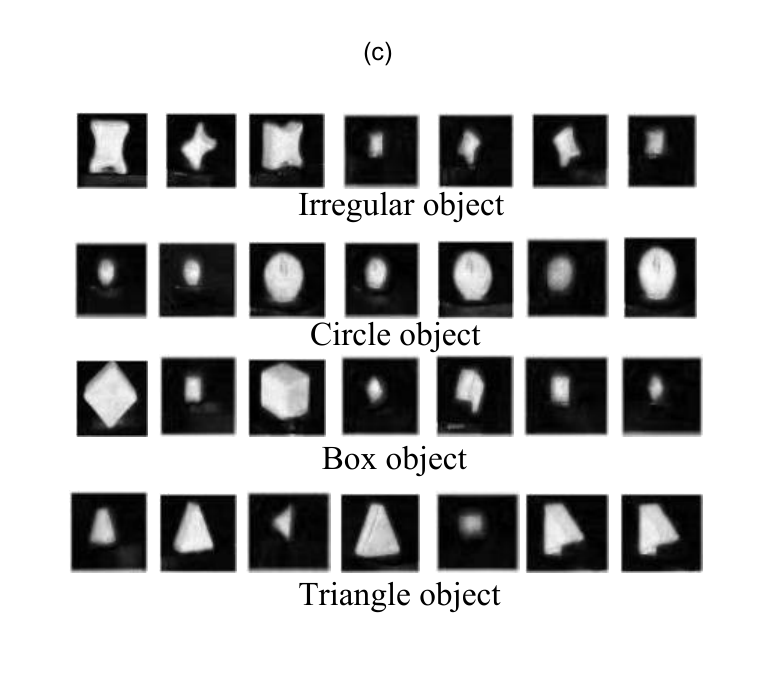}
\vspace{-1mm}
\caption{(a) Active classification accuracy for  recognition, (b) active classification error vs threshold $t_p(\hat{c})$ and (c) sample of extracted images used an inputs to the proposed HML-ELM.}\label{fig::classification_results_1}
\end{center}
\end{figure*}

From Table 3, it can be observed that the highest performance is obtained by a CNN. However, the computational load associated to its deployment is significantly higher to the rest of the techniques. On contrary, ML-ELM based architectures require a smaller computational load to provide a similar performance. Based on our experiments, for all models it was found the highest balance between object classification and low computational load can be obtained by using a value of $t_p(\hat{c}) \geq 0.8$. In Fig. \ref{fig::classification_results_1}(a) and (b), the confusion matrix obtained by the proposed HML-ELM and the average of the active classification error of five experiments with different values of $t_p(\hat{c})$ is illustrated respectively. In Fig. \ref{fig::classification_results_1}(a), it can be concluded that lowest performance achieved by the HML-ELM is for the recognition of irregular objects while the highest is obtained for the classification of rounded objects. In Fig. \ref{fig::classification_results_1}(b), the average classification error produced by the proposed HML-ELM for different values of $t_p(\hat{c})$ s illustrated. From Fig. \ref{fig::classification_results_1}(b), it can be also noted, a decreasing trend for the classification error is produced as the classification threshold increases. As pointed out in \cite{wagner2010toward}, it is known that the application of IT2 FLSs usually provides a more robust performance than their type-1 counterpart in the presence of uncertainty and noisy signals. This is mainly accredited to the ability of IT2 FLSs to better deal with uncertainty not only as a measure of ambiguity, but also  as a measure of deficiency that results from the boundary of imprecise fuzzy sets. As illustrated in Fig.  \ref{fig::classification_results_1}(c), although a process for image processing is implemented, a number of images may become noise-corrupted due to the orientation and size of the objects. In the HML-ELM, the implementation of an IT2 FLS with a SC for the classification of the extracted features results more practical than a type-1 fuzzy classifier for the treatment of noisy images with an improved model accuracy.

\begin{figure}
\begin{center}
\includegraphics[width=9.3cm,height=7.8cm]{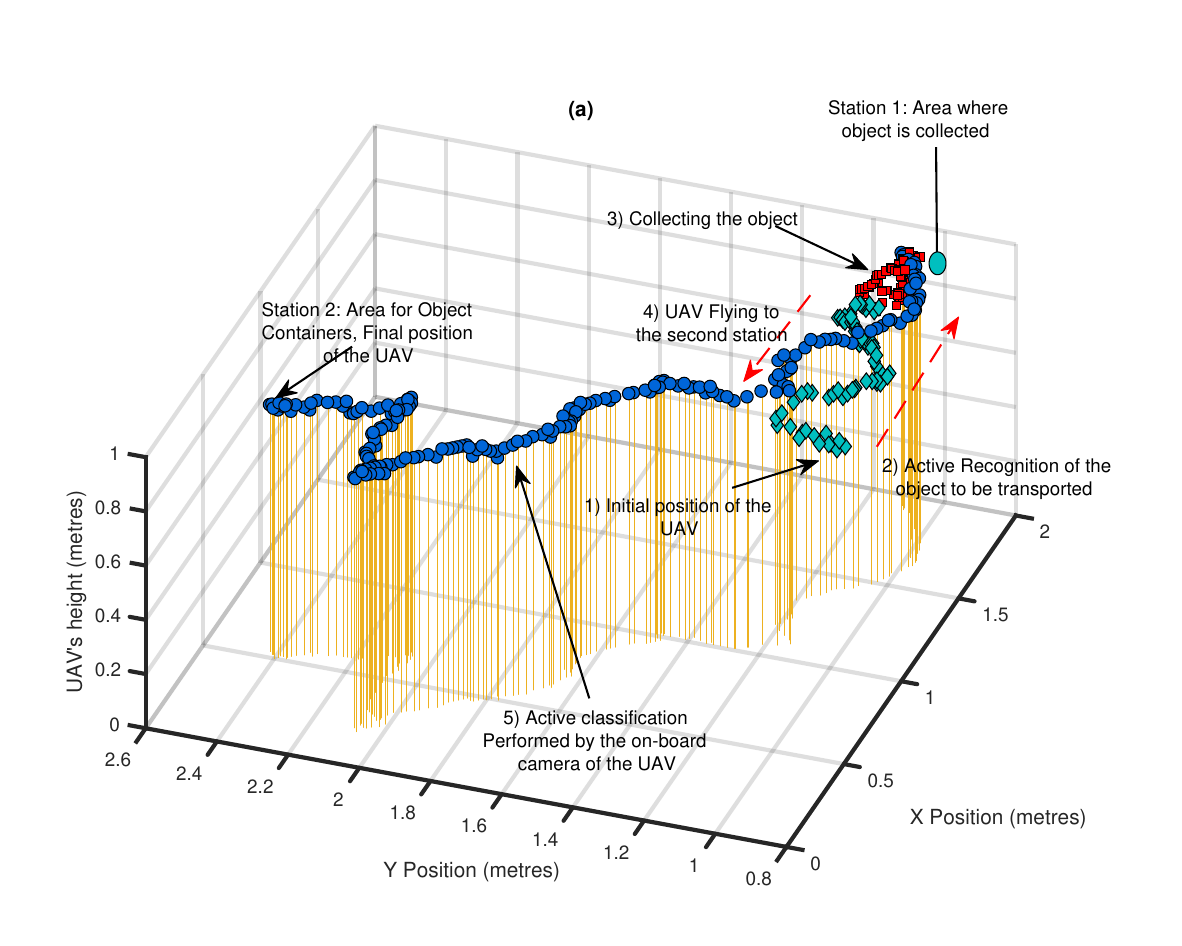}
\includegraphics[width=9.3cm,height=7.8cm]{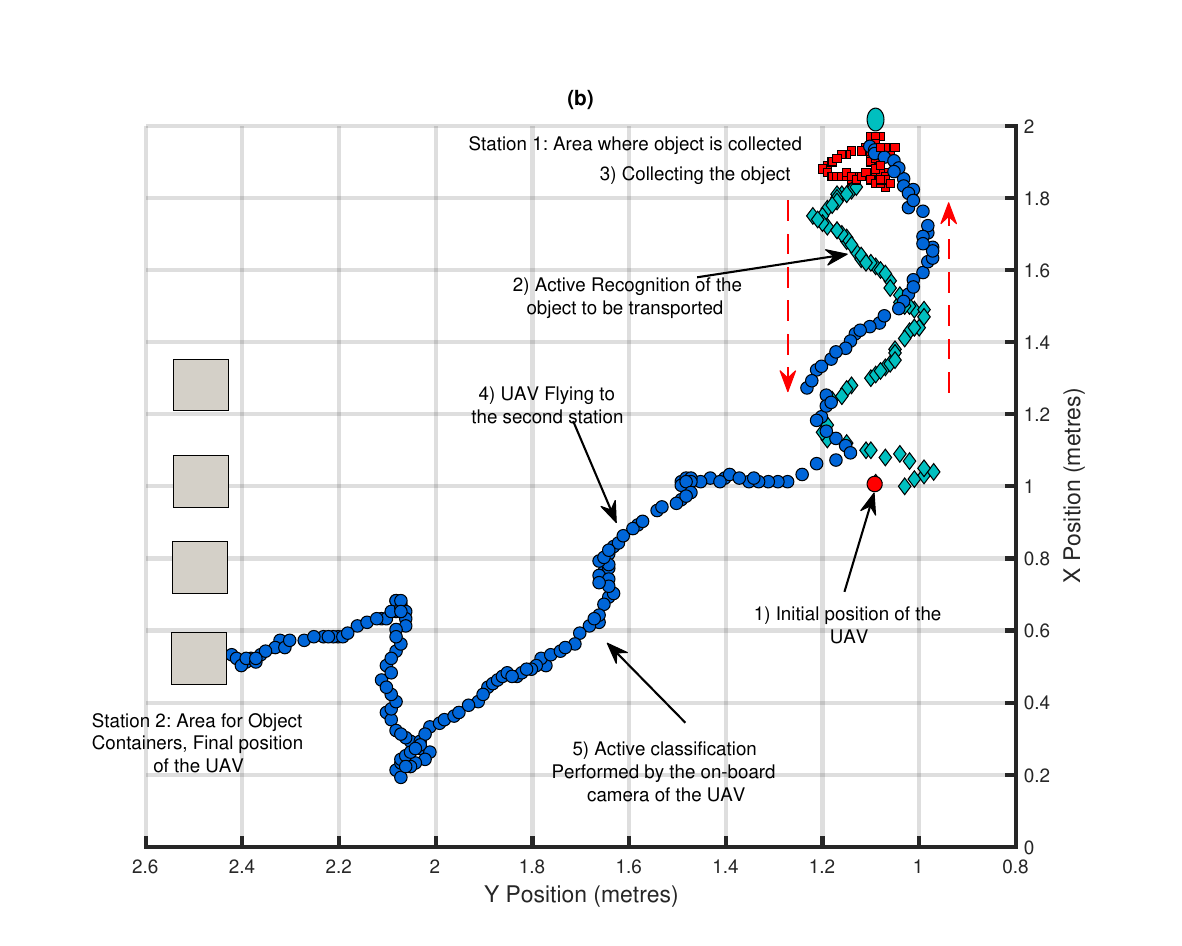}
\caption{Trajectory followed by the UAV in a random experiment in (a) space (3D) and (b) two-coordinates (2D).  2D.}\label{fig::drone_3D}
\end{center}
\end{figure}

In order to illustrate the trajectory followed by the UAV in a random experiment, in Fig. \ref{fig::drone_3D}, the position of the UAV is illustrated in 2D and 3D. From Fig. \ref{fig::drone_3D}(a), it can be observed the UAV is able to control its position in (X-Y-Z-axis(height) by using the information computed by the virtual sensors. This behaviour allows the UAV to autonomously decide when an object must be collected and then transport it to the final container. Compared to traditional motion planning algorithms based on GPS information, the proposed approach allows the UAV to navigate in indoors environments.    
\begin{tablehere}
\caption{\scriptsize Average performance of five trials for active object classification.}\label{iris_results} 
\centering 
\begin{tabular}{p{1.55cm} |p{0.9cm}|p{0.1cm} |p{0.1cm}| p{0.1cm} }
\hline 
 \scriptsize Models &\multicolumn{1}{c}{\scriptsize Image Classification time (Sec)}  & \multicolumn{3}{c}{\scriptsize Active classification ($\%$)} \\
\hline 
 \scriptsize  CNN & \multicolumn{1}{c}{\scriptsize 0.317} &\multicolumn{3}{c}{\scriptsize 98.62}  \\
 \scriptsize  ML-ELM & \multicolumn{1}{c}{\scriptsize 0.087} &\multicolumn{3}{c}{\scriptsize 93.10}  \\
 \scriptsize  ELM  & \multicolumn{1}{c}{\scriptsize 0.061} &\multicolumn{3}{c}{\scriptsize 69.01} \\
 \scriptsize  HML-ELM & \multicolumn{1}{c}{\scriptsize 0.098} &\multicolumn{3}{c}{\scriptsize 97.67} \\
 \scriptsize  ML-FELM & \multicolumn{1}{c}{\scriptsize 0.085} &\multicolumn{3}{c}{\scriptsize 97.30} \\
\hline
		
		\end{tabular}
		\centering 
		\label{nfm_results} 
\end{tablehere}
\vspace{3mm}
\section{Summary and Discussion}
In this study, a fast ML-ELM structure that unifies two different approaches based on the concept of ELM-AE and Simplified IT2 Fuzzy Extreme Learning Machine (SIT2-FELM) for the classification of images is suggested. The proposed methodology demonstrated to deliver a satisfactory performance compared to other existing ML-ELM approaches. From the comparative study presented in previous section, the following summary and discussion is provided:
\begin{itemize}
    \item[1.] The proposed HML-ELM is a hierarchical learning methodology that consists of two separate phases. First, a number of $L$ layers is used to perform unsupervised learning for feature data representation. Similar to traditional ML-ELM, each hidden layer for feature representation is an ELM-AE where input data is used as output $\textbf{x}=\textbf{t}$, random weights and random biases of the hidden nodes are chosen to be orthogonal. Activation functions in each ELM-AE can be either linear or nonlinear piecewise. In the second phase, the final features are classified using a new simplified IT2 Fuzzy Extreme Learning Machine  with a SC type reduction layer that we called SIT2-FELM for short. 
    \item[2.] Compared to deep learning frameworks trained with Back Propagation algorithms, each layer in the proposed HML-ELM is an independent block where fine tuning is not necessary. Hence, its  training can be much faster than traditional deep learning methods.
    \item[3.]  It is known that IT2 FLSs outperform their type-1 counterpart in a large number of applications. This is largely accredited to the capability of IT2 Fuzzy Sets (FSs) to better model the face of uncertainty and that each T2 FS has an uncountable number of embedded type-1 FSs. Therefore, by adding a classification layer of the proposed HML-ELM based on a SIT2-FELM, an improved performance for the classification of extracted features can be obtained.  
    \item[4.] In the design of IT2 FELMs, one of the most important steps is type reduction (TR). A frequent TR is the center of sets type reducer (COS TR) which requires the solution of two nonlinear constrained optimization problems. A common solution is to employ Karnik-Mendel algorithms (KM). The implementation of KM implies an iterative procedure to sort the upper and lower bounds of the parameters of the consequent parts of an IT2-FELM. This may represent a bottleneck in their deployment. In this study, by adding a SC algorithm as the type-reduction in an IT2-FELM, the need of sorting the consequent parts is eliminated. This is translated into a reduction in the associated computational load while providing similar model accuracy to that obtained by IT2-FELMs with a KM algorithm. 
\end{itemize}
For practical validation of the proposed HML-ELM in the field of UAVs, an efficient image classification mechanism has been suggested. The proposed HML-ELM is able to provide accurate information to the UAV to actively classify objects and navigate in indoor environments without the need of local GPS system. Within this context, it was also demonstrated that hybrid learning approaches from ELM theory and Interval Type-2 Fuzzy Logic (IT2 FL) for the classification of images can be deployed in real-world conditions. All these aspects permit the HML-ELM to be a suitable approach for the solution of complex classification problems where fast and robust decisions are required in the presence of noise and uncertainty. Moreover, by using a SIT2-FELM with a SC type reducer as the final layer of the HML-ELM, an improved performance compared to type-1 FLSs is assured.

The implementation of the HML-ELM in AUVs produced interesting results. First, it was possible to determine the optimal number of images necessary to accurately recognised an object from different positions and orientations. Secondly, by implementing the HML-ELM, the speed for decision-making was faster compared to other techniques such as CNN and ML-FELM. These features allow to have an efficient aerial robotic platform that intelligently decides when the object to be recognised needs to be collected and finally transported to the final goal. Additionally, the experiments demonstrated that the accuracy of the HML-ELM was not affected for real time classification. It is worth mentioning, the computational load grows linearly for very large image data sets. 
\section{Conclusions}

In this paper, a new Hybrid Multilayer Extreme Learning Machine (HML-ELM) that unifies the concept of ELM-based Autoencoder (ELM-AE) and Simplified Interval Type-2 Fuzzy ELM (IT2FELM) is suggested and applied for active image classification in the field of Unmanned Aerial Vehicles (UAVs).

The proposed HML-ELM is a hierarchical learning approach that involves two main phases to the training of multilayer neural structures. First, an unsupervised  step that consists of a number of ELM-AEs is used to extract a number of high-level feature representations. In the HML-ELM, each ELM-AE structure is a single-hidden layer Perceptron neural network. Secondly, the final features are classified by using a simplified version of Interval Type-2 Fuzzy Extreme Learning Machine (SIT2-FELM). The proposed SIT2-FELM incorporates as its type-reduction an improved version of the algorithm Center of Sets Type Reducer without Sorting Requirerment (COSTRWSR) called SC algorithm that eliminates the need of sorting the consequent parameters of an IT2-FELM with a Karnik-Mendel algorithm (KM). 

To evaluate the performance of the HLM-ELM, two different types of experiments were suggested. In the first experiment, the HML-ELM was applied to the classification of a number of benchmark data sets in the field of image classification as well as the object data set created in this study. Finally, the HML-ELM was implemented to solve a real-world problem in the field of unmanned aerial missions. This experiment involves the active recognition of four different objects and their collection and transport between two predefined locations. In the first location, an object is actively classified by the UAV by using a trained HML-ELM. Consequently, this information is used by the UAV to collect and transport the object to a second station. At the second location, active object classification is continuously required to identify the correct container at which the object needs to placed. The results show that the proposed approach can be efficiently implemented in real unmanned aerial missions where active classification of objects is required and used as a guidance mechanism for autonomous navigation. From our experiments, it can be concluded the proposed HML-MELM achieves a highe level representation and an improved image classification performance with respect to conventional ML-ELM, ELM and ML-FELM.


\bibliographystyle{unsrtnat}
\bibliography{references}  






\end{document}